%% file: CLEO_paper.tex
\documentclass[preprint,12pt]{elsarticle}

\usepackage{graphicx} 
\usepackage{subfigure}
\usepackage{float}

\usepackage{pseudo}
\usepackage[boxed,linesnumbered]{algorithm2e}
\providecommand{\DontPrintSemicolon}{\dontprintsemicolon}

\usepackage{array}
\usepackage{url}
\usepackage{hyperref}

\usepackage{amsmath,amssymb,amsfonts}
\usepackage{bm,bbm}

\usepackage{multirow}

\usepackage[inline,nomargin,draft]{fixme}
\usepackage{color}

\journal{Artificial Intelligence}
\input{definitions}

\begin{document}

\begin{frontmatter}



	\title{Learning Modulo Theories for preference elicitation in hybrid domains}
	

	\author[label1]{Paolo Campigotto\corref{cor1}}
	\ead{paolo.campigotto@tu-dortmund.de}
	\author[label2]{Roberto Battiti}
	\ead{battiti@disi.unitn.it}
	\author[label2]{Andrea Passerini}
	\ead{passerini@disi.unitn.it}

	\address[label1]{Informatik Lehrstuhl 4, 
					 Technische Universit\"{a}t Dortmund, \\
					 Otto-Hahn-Str. 16, D-44227 Dortmund, Germany.}

	\address[label2]{DISI - Dipartimento di Ingegneria e Scienza dell'Informazione, \\
					 Universit\`a degli Studi di Trento, \\
					 Via Sommarive 14, I-38123 Povo, TN, Italy.}

	\cortext[cor1]{Most of this work has been done while the author was with DISI.}
	
	\begin{abstract}
		This paper introduces CLEO, a novel preference elicitation algorithm capable of recommending complex objects in hybrid domains, characterized by both discrete and continuous attributes and constraints defined over them.		
		The algorithm assumes minimal initial information, i.e., a set of catalog attributes, and defines decisional features as logic formulae combining Boolean and algebraic constraints over the attributes.  
		The (unknown) utility of the decision maker (DM) is modelled as a weighted combination of features.
		CLEO iteratively alternates a preference elicitation step, where pairs of candidate solutions are selected 
		based on the current utility model, and a refinement step where the utility is refined by incorporating the feedback 
		received. The elicitation step leverages a Max-SMT solver to return optimal hybrid solutions according to the current
		utility model. The refinement step is implemented as learning to rank, and a sparsifying norm is used to favour the
		selection of few informative features in the combinatorial space of candidate decisional features.

		CLEO is the first preference elicitation algorithm capable of dealing with hybrid domains, thanks to the use of Max-SMT
		technology, while retaining uncertainty in the DM utility and noisy feedback. In so doing it adapts the recently introduced 
		learning modulo theory framework to the preference elicitation setting. The combinatorial formulation of the utility
		function coupled with the feature selection capabilities of 1-norm regularization allow to effectively deal with the uncertainty
		in the DM utility while retaining high expressiveness. Experimental results on complex recommendation tasks show the ability of 
		CLEO to quickly focus towards optimal solutions, as well as its capacity to recover from suboptimal initial choices. 
		While no competitors exist in the hybrid setting, CLEO outperforms a state-of-the-art Bayesian preference elicitation algorithm
		when applied to a purely discrete task.
	\end{abstract}

	\begin{keyword}
		preference elicitation \sep learning while optimizing
		\sep (Maximum) Satisfiability Modulo Theory \sep hybrid optimization. 

	\end{keyword}

\end{frontmatter}



\section{Introduction}
\label{sec:intro}

Automatically discovering the solution preferred by a decision maker (DM) from a large set of candidate ones is a key component of many systems, including decision-support, recommendation algorithms and personal agents. This task is usually referred to as the \emph{preference elicitation} problem~\cite{pe_survey08}.  
In principle, one may first ask the user to express her preferences and then translate them into a utility function defined over the search space of candidate solutions. The configuration maximizing the utility function is recommended to the DM. However, this approach is impractical, for several reasons~\cite{Bounded_ration_78}:
\begin{itemize}
	\item the user cannot usually define her preferences \emph{a priori}, without seeing any tentative results. Only when facing candidate solutions, she may realize ``what is possible'' and articulate her actual objectives;
	\item the cognitive effort and the time required to the user for completely specifying preferences are usually not affordable;
	\item in general, formalizing the user preferences as a mathematical model is not trivial: a model should capture the qualitative notion of preference and represent it as a quantitative function.
\end{itemize}
To handle the initial incomplete knowledge of the user utility, an incremental approach is usually adopted, where a configuration is recommended to the user based on \emph{partial} preference information only. If the user is not satisfied by the tentative solution, she is asked for additional preference information and a refined configuration is suggested. This incremental process needs techniques that can reason with partially-specified utility functions and take decisions under uncertain preference information. Furthermore, 
the interaction with human decision makers, with limited patience and bounded rationality, limits both the number and the complexity of the queries asked during the elicitation process, bounds the time needed for providing the recommendations and has to deal with inaccurate and inconsistent human feedback.  

The main requirements for practical applicability of preference elicitation are~\cite{GSMalgorithm10}:  
\begin{enumerate} 
	\item \emph{real-time} interaction with the DM, where both the query generation and the solutions recommendation must be accomplished in no more than few seconds; 
	\item \emph{robustness} to inconsistent and contradictory feedback from the DM characterizing the typical human decision making process; 
	\item \emph{cognitively affordable} queries to the user, i.e., comparison queries;
	\item \emph{scalable} methods, that evaluate at each preference elicitation stage a number of candidate queries that grows not more than linearly in the cardinality of the solutions space.
\end{enumerate}

Different approaches to preference elicitation have been proposed. 
Usually, a \emph{parametric} formulation of the space of possible DM utility functions is adopted. A set of basis functions are defined on subsets of the attributes, and the utility model is formulated as a weighted linear combination of these basis functions. 

Approaches to preference elicitation can be classified by the way they 
make recommendations under uncertainty in the weight values. 
Uncertainty in DM utility can be represented for instance by defining a space of \emph{feasible} weights, identified by bounds or constraints on the values. These constraints are learned from the preference information elicited from the DM.  
This popular approach, known in the literature as \emph{reasoning under strict uncertainty}, is adopted 
in~\cite{regretBased07,Boutilier2010,regretBased06}. In these papers, decisions under uncertainty are taken according to the \emph{minimax regret criterion}: the configuration minimizing the worst-case loss with respect to the feasible utility functions is recommended. 

The \emph{Bayesian approach}~\cite{gp2010,GSMalgorithm10,MC12,PL_multiple_DMs_12} maintains a probability distribution over the space of all possible weight values. Decisions are taken according to this probability distribution: the recommended solution is usually the one with greatest \emph{expected} utility. 

Recent work in the field of constraint programming~\cite{softConstr10} formalizes the user preferences in terms of \emph{soft constraints}. 
In soft constraints, a generalization of hard constraints, each assignment to the variables of one constraint is associated with a preference value. 
The work in~\cite{softConstr10} introduces a  preference elicitation strategy for soft constraint problems with missing preference values. 

However, neither the works~\cite{regretBased07,Boutilier2010,regretBased06} based on the minimax regret nor the constraint-based approach~\cite{softConstr10} can handle inaccurate and contradictory human feedback. 
The Bayesian method proposed in~\cite{GSMalgorithm10} satisfies all main requirements for practical applicability  discussed above. However, it can handle discrete attributes only, and it is hardly generalizable to the continuous case.

This paper introduces a novel algorithm which satisfies all the main principles for practical applicability of preference elicitation, 
allows to deal with hybrid domains and when applied to purely boolean problems consistently improves the state-of-the-art in terms of number of queries and quality of the returned solution.
The approach adopts a combinatorial formulation of the user utility function, modelled as a weighted combination of first-order logic formulae. Each formula combines predicates in a certain theory of interest by using the logical connectives. 
The theory fixes the interpretation of the symbols used in the predicates (e.g., the theory of arithmetic for dealing with integer or real numbers). 
For example, consider the case of flight selection. The predicate $\lmtphi_1 = (A_1+A_2 \leq 5 \mbox{ \emph {hours}})$ defines the preference for a travel duration, calculated as flight duration (continuous attribute $A_1$) plus transfer time to the departure airport ($A_2$), smaller than five hours. The predicate $\lmtphi_2 = ( A_3 < 2)$ states the desirability for a flight with a number of stopovers (discrete attribute $A_3$) smaller than two. 
The DM preferences about the candidate flights are expressed by associating the two predicates $\lmtphi_1$ and $\lmtphi_2$ with weights $w_1$ and $w_2$, respectively\footnote{In this simple example, each formula consists of a single predicate only. In the general case, arbitrary logic formulae (e.g., conjunctions or disjunctions of possibly negated  predicates) are considered. 
}. The flight 
maximizing the sum of the weights of the satisfied predicates is the one preferred by the DM. 

The configuration maximizing the weighted combinations of the first-order logic formulae is identified by applying a \emph{Maximum Satisfiability Modulo Theory} (Max-SMT) solver~\cite{NieOli06}. Max-SMT is a powerful recent formalism to optimize weighted formulae in a decidable first-order theory.  
Max-SMT enables to describe candidate solutions of the preference elicitation task by using both discrete and continuous attributes simultaneously (\emph{hybrid search domain}), thus improving the state-of-the-art of preference elicitation,  which cannot handle hybrid search domains. Furthermore, Max-SMT enables to manage complex non-linear interactions among the attributes (for example, a cost attribute defined as a function of the remaining attributes), increasing the expressiveness. \emph{Learning modulo theories} was recently introduced~\cite{Teso2015} as a framework
for adapting structured-output learning to hybrid domains by leveraging Max-SMT technology.
This paper adapts the framework to deal with preference elicitation tasks. 

The approach presented in this paper assumes a very limited amount of prior information about the task to be solved.
The initial knowledge is limited to a set of \emph{catalog attributes} used to describe the candidate solutions. 
The combinatorial formulation of the DM utility over the catalog attributes is initially unknown and needs to be learned by interacting with DM.
For this purpose, our approach consists of an iterative algorithm,
alternating a preference elicitation step guided by the currently
learned utility function and a refinement step where the quality of the utility
function is improved according to the feedback received.
In the preference elicitation step, two candidate configurations are selected according to the current utility and
presented to the DM for comparison.
The refinement step consists of solving a ranking problem which outputs a refined utility function consistent with the feedback received (soft consistency is allowed to deal with noisy feedback). The feature space of the utility function is given by all possible first-order logic formulae  combining the predicates up to a certain degree. 
Only a small fraction of these candidate features is actually part of the unknown utility for a certain DM~\cite{mil56}. A sparsifying norm \cite{Tib96}
is used during training in order to favour utility functions with few non-zero weights, thus performing
constraint selection in the combinatorial space of candidate features. 
In the rest of this
paper the algorithm is referred to by the acronym CLEO, which stands for
unknown Combinatorial utility
function joint LEarning and Otimization.

An experimental evaluation on realistic problems defined over hybrid
domains (i.e., with both discrete and continuous decisional attributes) and with inaccurate human feedback
demonstrates the effectiveness of CLEO in focusing towards the optimal
solutions, its robustness to noisy learning signals and its ability to recover from suboptimal
initial choices. While no competitors exist in the general case of
hybrid domains, we provide an experimental comparison on the
simplified task of learning purely Boolean combinatorial
functions. Thanks to its ability to learn complex non-linear
interactions between attributes, CLEO outperforms a state-of-the-art
Bayesian preference elicitation approach~\cite{GSMalgorithm10}.

A preliminary 
version of CLEO was presented in~\cite{stochLogicUtFun2010}. This manuscript extends it 
in a number of directions. First, it replaces quantitative
judgments asked to the DM with less
cognitive demanding queries, consisting of pairwise preferences of
candidate solutions. Second, it considerably extends the experimental
evaluation, including a more realistic recommendation problem. Third,
it provides a deeper comparison with the preference elicitation
literature, and adds an experimental comparison with a
state-of-the-art preference elicitation technique. 

The organization of the paper is as follows. Section~\ref{sec:backgr}
introduces the terminology and the notation used in the paper,
focusing in particular on the Max-SMT formalism. A small introductory
example of the preference elicitation tasks 
follows (Sec.~\ref{sec:example}).  The CLEO algorithm is introduced in
Sec.~\ref{sec:CLEO} and some of its main properties are analyzed in
Sec.~\ref{sec:CLEOanalysis}. Related work is discussed in
Sec.~\ref{sec:related}, while Section~\ref{sec:exp} reports the
experimental evaluation. Finally, a discussion including potential future 
research directions concludes the paper.


\section{Notation and background}
\label{sec:backgr}

This section provides the necessary background to introduce the CLEO algorithm. The Satisfiability Modulo Theory (SMT) formalism for solving decision problems over hybrid domains is explained, followed by its generalization (Max-SMT) to handle optimization tasks. 
Table~\ref{tab:notation} summarizes the notation used throughout the paper.

\begin{table}[H]
	\begin{tabular}{ll}
		\hline \hline
		{\bf Symbol}						 & {\bf Meaning}			\\
		\hline
		$\top$, $\bot$						 & Boolean values {\tt true} and {\tt false}		\\
		$x, y, z, \ldots$					 & Rational variables	\\
		$A_1, A_2, \ldots, A_n$				 & Catalog attributes (Boolean or rational variables)	\\
		$\lmta$						    	 & Configuration (assignment of values to all catalog \\
											 & attributes) \\
		$\lmta^i$							 & i-\emph{th} configuration \\
		$\lmtphi_1, \lmtphi_2, 
          \ldots, \lmtphi_m$  				 & Constraints. They can be atomic (Boolean attributes  \\
											 & or predicates over rational attributes, e.g. $x + y < 3$)\\
											 & or the combination of atomic constraints by the logical \\
											 & connectives (e.g. $\lnot has\_car \to dist\_supermarket \le \theta$)  \\
		$\mathbb{I}_k(\lmta) $ 				 & Indicator function for constraint $\lmtphi_k$ over $\lmta$. \\
											 & It evaluates to one if $\lmtphi_k$ is satisfied, to zero otherwise. \\
		$\lmtPSI(\lmta)$				     & Feature (i.e., constraint) representation of \\ 
											 & configuration $\lmta$ \\
		$\psi_k(\lmta) = \mathbb{I}_k(A) $   & Feature associated to constraint $\lmtphi_k$ \\
		\lmtw								 & Weights			\\
		\hline \hline
	\end{tabular}
	\caption{\label{tab:notation} Explanation of the notation used throughout the text.}
\end{table}

\subsection{Satisfiability Modulo Theory}
Propositional logic considers formulae involving Boolean variables and logical connectives.
The satisfiability (SAT) problem consists of deciding whether a formula in propositional logic 
can be satisfied by a truth value assignment of the Boolean variables. 
Satisfiability Modulo Theory (SMT)~\cite{BarSebSes09,sebastiani07} extends SAT to decide about
satisfiability of a first-order formula with respect to a {\em
  background theory} \T, like linear arithmetic over the rationals
(\larat{}) or integers (\laint{}), or a combination of theories. First-order logic involves variables, functions and predicates; the theory \T fixes the interpretation of predicate and function symbols.
For example, given the following SMT formula from the theory of arithmetic over integers:
$$ x+y+z \leq 4,\mbox{ } x,y,z \in \{1,2,3\}$$
we are interested in deciding whether there is an assignment of integer values to the variables $x$, $y$ and $z$ satisfying the formula. In this paper, SMT(\T) indicates satisfiability modulo theory \T, e.g., SMT(\larat{}) for satisfiability modulo linear arithmetic over the rationals.

Current SMT solvers are based on the so-called {\em lazy} approach,
where an outer SAT-solver interacts with one or more specialized
\T-solvers (one for each theory) in order to progressively focus the
search towards theory-consistent solutions or to state the unsatisfiability of the input SMT formula.
A \emph{T-solver} is a \emph{specialized} reasoning method for the theory \T integrated as submodule in the SMT solver.
Usually a T-solver is a decision procedure developed to check the satisfiability of \emph{conjunctions} of literals (i.e., atomic formulae and their negations) over theory \T. The generalization to arbitrary propositional structures is handled in conjunction with the SAT solver integrated in the SMT solver.
For ease of exposition, here a single theory is assumed, but all the machinery described can be applied to arbitrary
combinations of theories. 

Let $\Delta$ be an SMT formula made of predicates 
in a certain theory \T. Its {\em Boolean abstraction} $\Delta^-$ is
obtained replacing each i-\emph{th} theory-specific predicate in $\Delta$
with a Boolean variable $\lmtphi_i$, producing a formula in plain
propositional logic.  If this propositional formula in unsatisfiable,
the original formula $\Delta$ is also unsatisfiable and the whole SMT
solver stops.  Otherwise, the SAT solver finds a truth value
assignment to the Boolean variables $\lmtphi_i$ satisfying $\Delta^-$,
and presents it to the \T-solver to check for theory consistency.  The
\T-solver searches for an assignment of values to the theory variables
which is consistent with the solution provided by the SAT solver: if
the Boolean variable $\lmtphi_i$ is assigned a value true (false), the
corresponding i-\emph{th} predicate must (not) be satisfied by the values
assigned to the
theory variables. 
Predicates are evaluated using the rules of the theory \T.
If the \T-solver detects an inconsistency, it returns \unsat, plus a {\em
  justification}, i.e. a subset of the truth value assignment provided by the SAT solver which is
unsatisfiable according to the theory. The justification is an explanation of the inconsistency detected. 
This justification is added to the original formula, and the process is repeated until a
theory-consistent solution is found, or the refined formula is not
satisfiable.
 
\begin{example}
	Let $\Delta$ be the following SMT(\laint{}) formula:
	$$
		x+y+z \leq 3 \land (x \leq y \lor z=2) \land (x \geq 2 \lor x \neq z)
	$$
	where $x,y,z$ are integer-valued variables.
	Its Boolean abstraction $\Delta^-$ is: 
	$$
		\lmtphi_1 \land (\lmtphi_2 \lor \lmtphi_3) \land (\lmtphi_4 \lor \lmtphi_5).
	$$
	Suppose the SAT solver finds the following truth assignment satisfying $\Delta^-$:
	$$
		\lmtphi_1=\top, \lmtphi_2=\top, \lmtphi_3= \top, \lmtphi_4= \top, \lmtphi_5= \top.
	$$
	It corresponds to the following SMT(\laint{}) formula:
	$$
		x+y+z \leq 3 \land x \leq y \land  z=2 \land x \geq 2 \land x \ne z.
	$$ 
	When asked to evaluate this formula, the \T-solver detects that it is theory inconsistent, 
	since if $z$ is set to 2 and both $x$ and $y$ must be larger than 2, the sum of the three variables cannot be less than or equal to 3.
	A justification provided by the the \T-solver to explain the inconsistency may be, e.g., the following constraint: 
	$$
		\neg (\lmtphi_1 \land \lmtphi_2 \land \lmtphi_3 \land \lmtphi_4)
	$$
	which is included in $\Delta^-$ for the following calls to the SAT solver.
	A possible solution provided by the SAT solver for the refined Boolean abstraction: 
	$$
		\lmtphi_1 \land (\lmtphi_2 \lor \lmtphi_3) \land (\lmtphi_4 \lor \lmtphi_5) \land \neg (\lmtphi_1 \land \lmtphi_2 \land \lmtphi_3 \land \lmtphi_4)
	$$
	is the following truth assignment:
	$$
		\lmtphi_1= \top, \lmtphi_2= \bot, \lmtphi_3= \top, \lmtphi_4= \bot, \lmtphi_5= \top,
	$$
	corresponding to the theory formula:
 	$$
		x+y+z \leq 3 \land x > y \land z=2 \land x < 2 \land x \neq z.
	$$
	The \T-solver detects that this formula is theory consistent. It is satisfied, e.g., by the assignment:
	$$
		x=1, y=0, z=2.
	$$
	The search process of the overall SMT solver now stops, since a solution of the input formula $\Delta$ has been found.
\end{example}

These solvers are termed \emph{lazy} because of this incremental approach which generates constraints on demand, progressively refining the Boolean abstraction $\Delta^-$ by including additional theory-specific information. 

Modern lazy SMT solvers introduce a number of refinements to this
basic procedure, by pursuing a tighter integration between  SAT and  theory
solvers. A common approach consists of pruning the search space for the
SAT solver by calling the theory solver on partial assignments and
propagating its results. Furthermore, modern lazy SMT solvers combine solving techniques from very heterogeneous
domains. We refer the reader to~\cite{sebastiani07,BSST09HBSAT} for an
overview on lazy SMT solving.

\subsection{Max-SMT}
Max-SMT~\cite{nieuwenhuis_sat06,cimattifgss10,cgss_sat13_maxsmt}
generalizes SMT in the same way as Max-SAT does with SAT: rather than an assignment satisfying the input SMT formula, 
one maximizing the number of satisfied constraints is searched for. 
The weighted version of Max-SMT associates a (typically
positive) weight to each constraint, and the task is that of maximizing the weighted sum
of the satisfied constraints.

Let $\{(\lmtphi_1,w_1),\dots,(\lmtphi_m,w_m)\}$ be a set of
constraints with associated non-negative weights. The utility of any
assignment is clearly smaller or equal than the sum of all weights
$W=\sum_{i=1}^mw_i$ and larger than or equal to zero. The maximum-utility solution
is identified by a branch and bound strategy, which progressively tightens 
the upper and lower utility bounds and solves plain SMT problems encoding these bounds in their formulation. 
Given a lower bound $\hat{W} < W$, a solution is enforced to have a utility larger than $\hat{W}$ by generating a set of $m$ fresh Boolean variables and weights
$\{(\bar{\lmtphi}_1,\bar{w}_1),\dots,(\bar{\lmtphi}_m,\bar{w}_m)\}$ combined with the
following constraints~\cite{nieuwenhuis_sat06}:
\begin{align*}
\lmtphi_i \lor \bar{\lmtphi}_i &\quad  \forall i \in \{1,2, \dots ,m\} \\
\bar{\lmtphi}_i \to (\bar{w}_i = 0)  &\quad  \forall i \in \{1,2, \dots ,m\} \\
\lnot \bar{\lmtphi}_i \to (\bar{w}_i = w_i) &\quad \forall i \in \{1,2, \dots ,m\} \\
\sum_{i=1}^m\bar{w}_i \ge \hat{W}
\end{align*}
These constraints make any assignment with overall weight smaller than $\hat{W}$
inconsistent with the theory.


\section{An introductory example}
\label{sec:example}

Le us consider a customer that aims at building her own house. For this purpose, she asks a real-estate company about  potential housing locations. A very clear-headed person could formulate a request like:
\begin{quote}
{\it I would like a house in a safe area, close to my parents and to the
kindergarten, with a garden if there are no parks nearby. I would also like to live close to cycling and walking facilities. Of course, to fully enjoy these outdoor activities, the area should not be affected by air pollution. Finally, I prefer a site well served by public transport, with the nearest metro station easily reachable on foot. 
My maximum budget is 300,000 Euro.}
\end{quote}
These desiderata can be encoded as an SMT problem as follows:
\begin{align*}
  \textrm{solve:} & & \\
                  & (\lnot \lmtphi_1 \lor \lmtphi_2 ) \land \lmtphi_3 \land \lmtphi_4 \land \lmtphi_5 
					\land \lmtphi_6 \land \lmtphi_7 \land \lmtphi_8 \land \lmtphi_9  &\\
  \textrm{subject to:} & & \\
      & \lmtphi_1 =   A_2  					& \lmtphi_2 = A_1 				\\       
	  & \lmtphi_3 =  (A_3 < \theta_1)  		& \lmtphi_4 = (A_4 < \theta_2)	\\
	  & \lmtphi_5 =  (A_5 < \theta_3)       & \lmtphi_6 = A_6  	 		 	\\
	  & \lmtphi_7 =  (A_7 < \theta_4)  		& \lmtphi_8 = (A_8 < \theta_5)  \\
      & \lmtphi_9 =  (A_9 < \theta_6) 		&								\\ 
      & price(\lmta) \le  300000 &
\end{align*}
where the characteristics of the locations are defined by the set of catalog attributes $\lmta$ listed in Table~\ref{tab:housing_exa}. Function \textit{price} computes the price of location $\lmta$ based on the values of its attributes.
\begin{table}[h]
  \centering
\begin{tabular}{lll}
  name & description & type \\
  \hline
  $A_1$  & garden & Bool \\
  $A_2$  & park nearby & Bool \\
  $A_3$  & crime rate & Ordinal \\
  $A_4$  & distance from parents & Real \\
  $A_5$  & distance from kindergarten & Real \\ 
  $A_6$  & cycling and walking facilities in the neighborhood   & Bool   \\
  $A_7$  & air-pollution index &  Ordinal \\
  $A_8$  & public-transit service quality-index &  Ordinal \\
  $A_9$  & distance from nearest metro station  &  Real \\	
  $A_{10}$ & commercial facilities in the neighborhood & Bool \\
  $A_{11}$ & distance from downtown  & Real \\
\end{tabular}  
\caption{\label{tab:housing_exa}Catalog attributes for the housing example.}
\end{table}

If none of the locations available at the agency satisfies \emph{all} constraints,
the above problem has no solution.
A more reasonable alternative consists of solving the optimization version of the above problem, which maximizes the weighted sum of the satisfied constraints (i.e., a Max-SMT problem):
\begin{align*}
  \argmax_{\lmta} &  \quad \sum_{i=1}^5  \bar{w}_i & \\
  \textrm{subject to:} & & \\
	  & \lmtphi_1 =  (A_3 < \theta_1) & \\	
	  & \lmtphi_2 =  (\lnot A_2 \lor A_1) & \\
      & \lmtphi_3 =  ((A_4 < \theta_2) \land (A_5 < \theta_3)) & \\      
	  & \lmtphi_4 =  (A_6 \land (A_7 < \theta_4))  & \\
      & \lmtphi_5 =  ((A_8 < \theta_5) \land (A_9 < \theta_6))	& \\
      & price(\lmta) \le  300000 & \\
	  & \lmtphi_i \to (\bar{w}_i = w_i) \quad \forall i \in \{1,2, \dots ,5\} & \\
 	  & \lnot \lmtphi_i \to (\bar{w}_i = 0)  \quad  \forall i \in \{1,2, \dots ,5\} \\
\end{align*}
where each constraint $\lmtphi_i$ is associated to a weight $w_i$ 
quantifying the (relative) utility of the constraint.  The bound on
the price is a hard constraint that needs to be satisfied, thus it has
no weight. 

A fully specified scenario like the one described here is however not
realistic when a human DM is involved. An exact
specification of the set of relevant constraints is hard to obtain,
let alone their respective weights. The most natural scenario
consists of an interactive process, with the customer evaluating
candidate locations and the realtor updating her understanding of the
customer preferences according to the feedback received. The rest of
this paper introduces the CLEO algorithm, a preference elicitation
method that automatizes this process. 

Let us finally note that not all the catalog attributes describing candidate house locations 
may be relevant for a customer: in the above example the customer decides without considering  
the last two attributes in Table~\ref{tab:housing_exa}. 
A large list of catalog attributes enables both a fine-grained description of the locations 
and the interaction with different classes of customers, having different decisional items.
On the other hand, users are expected to take decisions based on a limited set of attributes 
in the large catalogue. The CLEO algorithm can identify 
the subset of catalog attributes relevant for a certain customer.


\section{The CLEO algorithm}
\label{sec:CLEO}

This section introduces the CLEO algorithm, first describing its components and then combining them
into the overall algorithm.

\paragraph{{\bf Catalog attributes}} CLEO assumes a {\it catalogue} of attributes 
which can be used to describe the configurations. Each configuration is an instantiation of the catalog attributes.  These attributes can be either Boolean (e.g., {\tt there is a garden}), ordinal (e.g., {\tt crime rate}) or real (e.g.,  {\tt distance to kindergarten}) variables (see Table~\ref{tab:housing_exa} in the previous example for a list). 
A large number of attributes can be included, in order to increase the expressiveness of the method and enable fine-grained descriptions of the configurations.
However, only a limited subset of the attributes may be relevant for a specific decision maker, and, in general, the subset varies when different users are considered. 
This section will show how CLEO identifies the subset of relevant attributes. 

\paragraph{{\bf Hard constraints}} Some combinations of attribute values may be infeasible. For example, in the above housing example, house locations with cost value smaller than a given threshold may not be available.
Arbitrarily-complex hard constraints define the \emph{feasible search space} of candidate configurations. The hard constraints are assumed to be known in advance. The CLEO algorithm provides to the DM \emph{only feasible solutions} during the preference elicitation process. 

\paragraph{{\bf Soft constraints}} \emph{Soft constraints} are defined over the catalog attributes. 
A soft constraint may or may not be satisfied by a feasible configuration. 
Each soft constraint is associated with a weight, defining the utility value of the constraint. \emph{Positive} weights are associated with constraints expressing positive preferences of the DM (i.e., features that the preferred configuration should have), while \emph{negative} weights are associated with constraints articulating negative preferences (i.e., features that the ideal configuration should \emph{not} have). 
The absolute value of the weight defines how much the soft constraint is relevant for the DM (w.r.t. to the other soft constraints). A \emph{zero}-weight identifies a constraint not considered by the DM. 

\paragraph{{\bf Space of soft constraints}} Soft constraints are
atomic constraints or their combination.  {\it Atomic} constraints are
constructed from catalog attributes, by simply taking their values for
Boolean variables, and constraining each ordinal and real variable to
be below a certain (variable-specific) threshold. In the case of non-Boolean variables,
atomic constraints are thus predicates in first-order logic.  More
complex constraints can be constructed by arbitrary combinations of
these building blocks. For example, {\tt distance to kindergarten
  $< \theta$} $\land$ {\tt distance to parents $< \theta$}, so that a
car is not needed, or, {\tt house with garden} $\lor$ {\tt distance
  from nearest park $< \theta$}, so that open-air activities are
possible.

These combinations are arbitrary logic formulae (e.g, conjunctions or disjunctions) of up to $d$ atomic constraints. 
The maximal degree $d$ contributes to limit the size of the soft constraints space, and is
grounded on the bounded rationality of humans, who can simultaneously
handle only a limited number of features.

The space of constructible soft constraints is clearly
exponential in the size of the catalogue. 
In the following we show how CLEO manages the large dimensionality of the soft constraints space.
For this purpose, let us define here the mapping function $\lmtPSI(\lmta)$ which projects configuration $\lmta$ into the space of all possible soft constraints, i.e., combinations of up to $d$ atomic constraints. 
Each soft constraint $\lmtphi_k$ is associated with its indicator function $\mathbb{I}_k(\lmta)$ which evaluates to one if the constraint is satisfied and to zero otherwise. 
The feature (i.e., constraint) representation of configuration $\lmta$ is the vector obtained by concatenating the evaluation of each indicator function:
$$
\lmtPSI(\lmta) = (\mathbb{I}_1(\lmta), \mathbb{I}_2(\lmta), \dots, \mathbb{I}_m(\lmta))
$$ 
In the following, the vector returned by function $\lmtPSI$ and the space of all possible vectors returned by function $\lmtPSI$ will be referred to as \emph{feature vector} and \emph{feature space}, respectively. The terms \emph{feature} and \emph{constraint} will thus be used interchangeably.

\paragraph{{\bf Combinatorial utility function}} 
The DM utility function is represented by a subset of the soft constraints defined over the catalog attributes. The soft constraints involved in the definition of the utility function are associated a weight different from zero and encode the DM preferences. 
The utility of a configuration is the sum of the weights of the soft constraints satisfied by the configuration. 

The above introduced feature vector $\lmtPSI(\lmta)$ enables the following compact formulation of utility function $f$:
\be
	\label{eq:ut}
	f(\lmta)=\lmtw^T \lmtPSI(\lmta)
\ee
where the weight vector $\lmtw$ contains the weights associated with
the candidate soft constraints. Due their bounded rationality and
limited information-processing capabilities, humans can 
handle only a limited number of features to make decisions. Thus only
\emph{very few} of the candidate soft constraints will actually be
considered by the DM, resulting in an extremely sparse weight vector
$\lmtw$.  This {\em sparsity assumption} will be accounted for when introducing the
learning stage.

\paragraph{{\bf Learning phase}} Learning amounts to find the
weights for the utility function formulation in Eq.~\ref{eq:ut} matching the unknown DM
preferences. Training examples for this phase consist of candidate
configurations with their evaluation from the DM.  
Asking quantitative feedback such as
real-valued scores is typically not affordable for a human DM~\cite{GSMalgorithm10}. A more
realistic scenario consists of asking the DM to rank solutions by
preference. We can thus formulate the problem as {\em learning to
  rank}, where the task is learning a function returning the same
ranking as the one provided by the DM. We focus on the
adaptation of SVM for ranking~\cite{ranksvm}, which assumes pairwise
ranking preferences, and enforces a (soft) large margin between the
two predictions. 
However, we have an additional requirement, which is
the sparsity assumption in the weight vector $\lmtw$.  Indeed, the feature vector
contains all possible constraints (up to a certain complexity), and
the learning phase should also perform some form of {\it constraint
  learning} by selecting a small set of relevant ones. 
Feature selection is in fact crucial to maximize
the learning accuracy with data sets characterized by redundant and
irrelevant features~\cite{Friedman04}.
We favour 
feature selection by replacing the 2-norm of SVM with a 1-norm,
which is a sparsifying norm encouraging solutions with few non-zero
weights~\cite{Friedman04}. The resulting learning problem is:
\begin{eqnarray}
\label{eq:svranking}
\min_{\lmtw,\lmtxi \geq 0} & & ||\lmtw||_1 + C \sum_{\lmta^i \succ \; \lmta^j}\xi_{i,j}^2 \\
\textrm{subject to:} & & \lmtw^T(\lmtPSI(\lmta^i)-\lmtPSI(\lmta^j)) \ge 1 - \xi_{i,j} \nonumber \\
                     & & \forall \; \lmta^i \succ \lmta^j \in \calD \nonumber
\end{eqnarray}
where $\lmta^i \succ \lmta^j$ indicates that configuration $\lmta^i$
is ranked before $\lmta^j$ in the DM preference. Constraints enforce
pairwise rankings to match DM preferences. A quadratic penalty
$ \xi_{i,j}^2$ is added to the objective function when a less preferred
solution gets a utility score which is not sufficiently smaller than
the more preferred one.
The regularization
parameter $C$ trades-off matching DM preferences with sparsity of the
weight vector, and is optimized during the learning process as
discussed further down.

\paragraph{{\bf Optimization phase}} The ultimate goal of
the algorithm is returning the best possible instance given the DM 
utility function.  However, since the utility function is
unknown, a preference elicitation phase is needed to 
gather information on DM preference and use it to refine the current
approximation $\hat{f}$ of her utility. 
CLEO asks the DM for pairwise comparisons of configurations.
The two configurations to be compared by the DM are generated by optimizing the learned 
utility function $\hat{f}(\lmta)$ twice. 
Since the learned utility function is a weighted combination of soft constraints involving Boolean variables and first-order logic predicates defined over discrete and continuous variables, it is optimized by using an off-the-shelf Max-SMT solver, which can efficiently reason in these hybrid domains.
The two optimization runs are performed based on  
the following principles:
\begin{enumerate}
\item the generation of top-quality configurations, consistent with
  the learned DM preferences;
\item the generation of diversified configurations, i.e., alternative
  possibly suboptimal configurations with respect to the learned
  utility $\hat{f}$;
\item the search for catalog attributes relevant to the DM not recovered by the
  current approximation $\hat{f}$, i.e., attributes not appearing in any
  of the soft constraints in $\hat{f}$.
\end{enumerate}

The rationale for the first principle is focusing on the relevant
areas of the utility surface, those of interest to the DM. As a matter
of fact, a preference elicitation system that asks to rank low quality
configurations
will be likely considered useless or annoying by
the 
DM~\cite{GSMalgorithm10}. 
In addition, the goal of CLEO is the identification of the solution preferred by the user (\emph{learning to optimize}) rather than an accurate global approximation of the DM utility function 
(\emph{learning per se}). 
This requires a shift of paradigm with respect to standard machine learning strategies, in order to model the relevant areas of the optimization fitness surface rather than reconstruct it entirely.

The second principle advocates the introduction of some diversification in the search,
by exploring the neighbourhood of the best solution for the currently learned preference
model $\hat{f}$. Finally, as the learned formulation
of $\hat{f}$ may miss some of the user decisional attributes, their search
is explicitly promoted by the third principle.
The need for a set of good 
and diverse configurations to be evaluated by the user is suggested also in~\cite{PuChen2008}.

Our optimization phase works as
follows.  First, $\hat{f}$ is maximized (first principle), generating
the first candidate configuration $\lmta^*$. Then, a {\it hard}
constraint is added to the Max-SMT problem as the disjunction of all
soft constraints not satisfied by $\lmta^*$, and maximization is run
again. This accounts for the second principle, by enforcing a new
solution $\lmta^{**}$ which differs from $\lmta^*$ by at least one
soft-constraint.  
If $\lmta^*$ satisfies all soft constraints in $\hat{f}$, the additional \textit{hard} constraint
generated is: $(\lnot A_1^* \lor \lnot A_2^* \ldots \lor \lnot A_n^*)$ 
which excludes $\lmta^*$ from the set of feasible solutions. 

Finally, each unassigned attribute, i.e., catalog attribute not appearing in any hard constraint or soft constraint with non-zero weight, in both $\lmta^*$ and
$\lmta^{**}$ is given a random value in its domain,
thus incorporating the
third principle. Indeed, if these catalog attributes are truly irrelevant for
the DM, setting them at random should not affect the evaluation of the
candidate solutions. On the other hand, if some of them are needed to
explain the DM preferences, driving their elicitation can allow to
identify the deficiencies of the current approximation $\hat{f}$ and
recover previously discarded relevant decisional items.


\begin{algorithm}[t!]
  \DontPrintSemicolon
	\caption{\label{alg:lmt4pref} Pseudocode for CLEO.}
	\begin{small}
	\KwData{Set of catalog attributes, set of atomic constraints, set of hard constraints}
	\KwResult{Most preferred solution $\lmta^*$}
        \tcc{Initialization}
        Select three configurations uniformly at random \;
        $ \calD \leftarrow \textrm{ranking of configurations by DM}$ \;
        \tcc{Refinement}
	\While{true}{
          \tcc{learning}
          \begin{eqnarray*}
            \hat{f}       & \leftarrow	& \argmin_{\lmtw,\lmtxi \geq 0} ||\lmtw||_1 + C \sum_{\lmta^i \succ \; \lmta^j}\xi_{i,j}^2 \\
                          & \text{s.t.} & \lmtw^T(\lmtPSI(\lmta^i)-\lmtPSI(\lmta^j)) \ge 1 - \xi_{i,j} \\
                          &             & \forall \; \lmta^i \succ \lmta^j \in \calD
          \end{eqnarray*} \;   

		  \tcc{optimization} 
		  Recommend configuration $\lmta^* = \argmax \hat{f}$  to the DM \;
          \If{ termination criterion is not satisfied}{
			\tcc{preference elicitation}  
		  	Generate $\lmta^{**}$ by diversification strategy \;
			$ \calD \leftarrow \calD \cup \textrm{ranking of pair} \mbox{ } (\lmta^*, \lmta^{**})\mbox{ } \textrm{by DM}$ \;
			 }
        }
	\end{small}
\end{algorithm}

\paragraph{{\bf Overall algorithm}} The pseudocode of the full
CLEO algorithm is shown in Algorithm~\ref{alg:lmt4pref}. It takes as input
the set of catalog attributes, the set of atomic constraints, the set of hard constraints defining the feasible configurations, and returns the solution which is most preferred by the DM. 
In the initialization
phase, 
the DM is asked for two pairwise comparisons of configurations selected by CLEO independently and uniformly at random in the feasible search space. 
Then a refinement
loop begins, where at each iteration first an approximation of the DM utility function is learned
using the current feedback. The refinement amounts at solving the ``learning to rank'' problem in
Eq.~(\ref{eq:svranking}), where $\calD$ is the dataset of all
pairwise preferences collected so far. The regularization parameter
$C$ is set to one in the first iteration, and fine-tuned by an
internal cross validation on the training set in the following ones.
With a slight abuse of notation, we write $\hat{f} \gets \argmax$ to
indicate that $\hat{f}$ is the function whose weights $\lmtw$ are the
result of the maximization. 
The configuration $\lmta^*$ maximizing the learned utility function $\hat{f}$   
is recommended to the DM.
If she is not satisfied with the suggested solution, 
an additional optimizer $\lmta^{**}$ of $\hat{f}$ is generated, favouring diversity between 
$\lmta^*$ and $\lmta^{**}$ based on the diversification strategy defined above. 
The dataset $\calD$ is then updated by including the comparison between $\lmta^*$ and $\lmta^{**}$ 
performed by the DM.

Being an interactive process involving a human DM, the most
obvious termination condition is the DM satisfaction with the current
recommendation. Additional conditions could be conceived, for
instance, by estimating the improvement one could expect by further
refining the utility function. We will discuss this and other
potential extensions in the conclusions.



\section{CLEO properties}
\label{sec:CLEOanalysis}

The CLEO algorithm has \emph{no free parameters} to be manually tuned. The number of  
iterations does not need to be fixed at the beginning. The DM may ask for an additional iteration by comparing the recommended configuration $\lmta^*$ with her own preferences. The termination criterion is thus represented by the satisfaction of the DM with $\lmta^*$. The regularization parameter $C$ in Eq.~(\ref{eq:svranking}) is set to one in the first iteration, and fine-tuned by internal cross-validation on the training set in the following ones.
In the first iteration, two pairwise comparisons are asked to the DM, while in the following iterations a single pairwise comparison is asked. The configurations to be compared at the first iteration are generated by sampling independently and uniformly at random the feasible search space. The evaluation of diverse examples stimulates the preference expression, especially when the user is still uncertain about her final preference~\cite{PuChen2008}. In particular, the diversity of the proposed solutions helps the user to reveal the hidden preferences: in many cases the decision maker is not aware of all preferences until she sees them violated. For example, a user does not usually think about the preference for an intermediate airport until a solution suggests an airplane change in a place she dislikes~\cite{PuChen2008}. 

The human cognitive capabilities bound the number of catalog attributes and the size $d$ of soft constraints. The limited size of the Max-SMT instances generated by CLEO enables the systematic investigation of the search space by means of a complete solver, which ensures the identification of a global maximum $\lmta^*$ of the learned utility model $\hat{f}$ (completeness property). However, CLEO cannot guarantee the quality of the model $\hat{f}$ approximating the true DM utilities, and therefore the optimality of $\lmta^*$ (or bounds on its quality) w.r.t. the \emph{true} DM utilities cannot be proved. As a matter of fact, the learning task in Eq.~(\ref{eq:svranking}) is convex, and thus guaranteed to converge to its global optimum, but the consistency of the learning algorithm with the true underlying user utility is only guaranteed asymptotically (i.e., provided that enough training data is available).
On the other hand, CLEO does not need to learn the exact form of the DM utility function. The goal of
our approach is indeed to elicit as few preference information from
the DM as possible in order to identify her favourite solution
(\emph{learning to optimize}).  For example, consider the toy DM
utility function represented by the negation of a single ternary term:
$\neg (\lmtphi_1 \land \lmtphi_2 \land \lmtphi_3)$.  The approximation of the DM utility
function consisting of the formula $\neg \lmtphi_1$ is sufficient to
find one of the favourite DM solutions. More in general, only the shape of
the utility function locally guiding the search to the correct
direction is actually needed. Indeed the experimental results reported
in Sec.~\ref{sec:exp} show the ability of CLEO in identifying
the optimal solution and the improvements in the quality of the
candidate solutions when increasing the number of refinement iterations (anytime
property).

Finally, CLEO satisfies the main requirements for practical applicability of 
preference elicitation. 
In detail:  
\begin{enumerate}

	\item \emph{multi-attribute} models. Candidate configurations are described by multiple decisional attributes. Since these attributes usually vary with different decision makers, CLEO assumes a set of catalog attributes, from which the  decisional items of a specific DM are automatically selected. Unlike the state-of-the-art methods (see Sec.~\ref{sec:related}) for preference elicitation, CLEO can handle both discrete and continuous-valued attributes simultaneously, thanks to the Max-SMT formalism which can efficiently tackle hybrid domains;

	\item \emph{real-time} interaction with the DM. 
Due to the limited number $n$ of catalog attributes and to the bounded size $d$ of soft constraints, the learning phase (problem~(\ref{eq:svranking}))
is accomplished in a negligible amount of time (w.r.t. the user
response time). An analogous observation holds for the computational
effort required by the optimization phase. Proposing a query consists of
generating two candidates to be compared.  Each candidate is obtained by
a run of the complete Max-SMT solver. The bounded value of $n$ and the efficient performance of
modern SMT solvers, that can efficiently manage problems with thousands of
variables and millions of constraints, enable the completion of the optimization phase in a negligible amount of time;
 
	\item \emph{robustness} to inconsistent and contradictory human feedback. The adoption of regularized machine learning strategies in CLEO enables a robust approach that can handle inaccurate (pairwise) comparisons of solutions from the DM. Assuming that a user always provides accurate and consistent preference
information is not realistic. 
Different factors may generate uncertain and inconsistent
feedback from the DM, including occasional inattention, embarrassment
when comparing very similar solutions or solutions which are very
different from her favourite one, DM fatigue increasing with the
number of queries answered;
 
	\item \emph{user cognitive load}. CLEO asks the user just for pairwise comparisons of candidate solutions. 
Most users are typically more confident in comparing solutions, providing
qualitative judgments like ``I prefer solution $\lmta^*$ to
solution $\lmta^{**} \mbox{ }$", rather than in
specifying how much they prefer $\lmta^*$ over $\lmta^{**}$;

	\item \emph{scalability}. At each preference elicitation stage, just one candidate query is considered by CLEO, independently of the cardinality of the configuration space. 
The adoption of 1-norm regularization for the formulation of the learning problem requires that the input catalog  attributes are explicitly projected in the feature space, i.e., the space of all possible soft constraints. Dealing with the explicit projection $\Phi$ in Eq.~(\ref{eq:svranking}) is tractable only for a
rather limited number of catalog attributes and size of constraints $d$. 
However, this will typically be the case when interacting with a human DM. Research in psychology has indeed shown that humans cannot handle simultaneously more than few ($7 \pm 2$) factors~\cite{mil56}. 

\end{enumerate}


\section{Related work} 
\label{sec:related}

The problem of automatically learning utility functions and eliciting
preferences is widely studied within the Artificial Intelligence community~\cite{Braziunas06techre,PEoverview11}. 
Different approaches have been proposed to take decisions with partial preference information during the elicitation process. The uncertainty in the utility function is usually represented by a set of feasible
utility functions (reasoning under strict uncertainty)~\cite{regretBased07,Boutilier2010,regretBased06}, which is narrowed down when additional preference information is elicited, or by a probability distribution over possible utility functions (Bayesian approach)~\cite{gp2010,GSMalgorithm10,MC12,PL_multiple_DMs_12}, refined when 
additional knowledge of the DM preferences is obtained. 
Finally, a recent line of research~\cite{softConstr10} developed within the Constraint Programming community shares with CLEO the combinatorial formulation of the DM utility function (constraint-based preference elicitation). In the following, these approaches to preference elicitation are reviewed and compared with CLEO. We also motivate the choice of the Bayesian method introduced by Guo and Sanner in~\cite{GSMalgorithm10} as benchmarking algorithm in our experiments and summarize its main features.
A more detailed description and discussion about the state-of-the-art methods for preference elicitation can be found in~\ref{app:sta_detailed}. 
 

\subsection{Strict uncertainty} 

A popular approach to model the uncertain knowledge about the DM
preferences consists of assuming a set of hypotheses, with no belief
on their strength. The set of hypotheses contains the feasible utility
functions and reflects the partial knowledge about the DM preferences.
The uncertainty about the DM preferences is decreased by restricting
the feasible hypothesis set, when relevant preference information is
received during the elicitation
process. 
This approach is often referred to as \emph{reasoning under strict
  uncertainty}~\cite{Braziunas06techre}.
  
The \emph{minimax regret} criterion~\cite{Savage1951} from statistical decision theory
provides a way to make decisions under uncertainty. Given a certain decision $\lmta$, the maximum regret is
the difference in utility between the DM most preferred solution
$\lmta^*$ and $\lmta$ assuming the worst-case scenario, where the DM
utility is the one in the feasible set for which this difference is
maximal. By adopting the minimax regret criterion, the decision
that minimizes this regret is taken. This criterion therefore suggests a robust 
decision w.r.t. the worst possible case. 
The recent work in~\cite{regretBased07,Boutilier2010,regretBased06} introduces
an approach to preference elicitation based on the minimax regret criterion. 
Queries to be asked to the DM are selected so as to reduce the minimax regret by
restricting the feasible hypothesis set. An advantage of minimax
regret approaches with respect to our formulation is that they can
provide theoretical guarantees in terms of bounds on the solution
quality and convergence to provably-optimal results. On the other
hand, these approaches assume perfect feedback from the DM and cannot handle the imprecise and contradictory information which is typical of interactions with human DM. Therefore, they are not suitable for the realistic preference elicitation tasks considered in this work.


\subsection{Bayesian uncertainty}
\label{subsec:Bayesian}

An alternative uncertainty model (Bayesian approaches) consists of defining a \emph{probability distribution} (or belief) over the candidate utility functions~\cite{gp2010,GSMalgorithm10,MC12,PL_multiple_DMs_12}. 
The probabilistic framework offers a flexible approach to preference elicitation, 
 handling the uncertainty in both utility and DM feedback. 
The \emph{expected utility} of a configuration is defined as the average utility computed with respect to the probability distribution over the utility functions. The configuration maximizing the expected utility is usually recommended to the user. Therefore, under the Bayesian paradigm, robust decisions are taken to minimize risk in expectation.
Queries are asked to the DM in order to increase the posterior probability of her utility.
The probabilistic framework enables to estimate the informativeness of the candidate queries. 
At each stage of the preference elicitation process the maximally
informative query is asked. The maximum expected loss (MEL) of taking a decision
$\lmta$ is the maximum expected reduction in utility when choosing
$\lmta$ instead of the DM most preferred solution $\lmta^*$, where
expectation is taken over the probability distribution of the utility
functions. The {\em value of information} (VOI) criterion suggests the query generating the largest expected
reduction in MEL. Exact computation of VOI, as well as exact
computation of the posterior distribution over utility functions given
the feedback, are extremely expensive. The state-of-the-art approaches~\cite{gp2010,GSMalgorithm10} resort to approximate solutions. 

The closest approach to CLEO is the Bayesian method introduced by Guo and Sanner in~\cite{GSMalgorithm10} (referred to as GSM). Indeed, unlike the techniques based on minimax regret and on the constraint satisfaction formalism, CLEO and GSM satisfy all the main principles~\cite{GSMalgorithm10} needed for practical applicability of preference elicitation (see Sec.~\ref{sec:intro}).


The GSM algorithm~\cite{GSMalgorithm10} 
searches for the configuration preferred by the DM within a given set of candidates.
The configurations are described by $n$ \emph{discrete} attributes $x_1, \dots, x_n$, where 
the k\textit{-th} attribute is assigned values from a finite set $X_k$ with cardinality $|X_k|$. 
The user utility functions are represented by a weight vector $\w$ with dimension $ \sum_{k=1}^n |X_k| $ specifying the utility of \emph{each} attribute value in $X_k$ for \emph{each} attribute $k$. This modelling choice assumes \emph{preferential independence} among the set of attributes. 



The uncertainty about the user preferences is represented by considering the weight vector $\w$ as a multivariate continuous random variable and by maintaining a probability distribution $\Pr(\w)$, which is incrementally refined. 
Different strategies are defined to select the query to be asked at each refinement stage. 
Since GSM asks pairwise comparisons to the DM, 
in principle the VOI of each possible pairwise comparison has to be estimated. This query strategy, termed \emph{informed VOI}, thus scales quadratically with the number of configurations and 
its computational cost is affordable for small search spaces only. 
In the experiments reported in~\cite{GSMalgorithm10}, already 20 configurations prevent its application, even if the probabilities of the two possible answers to a pairwise comparison are assigned fixed arbitrary values (\emph{uninformed VOI} strategy) rather than the values estimated from the elicited preference information. The computational load can be decreased by restricting the set of candidate pairwise comparisons, e.g., by fixing one  element of each candidate pair to the configuration $\x^*$ with greatest expected utility (\emph{restricted informed VOI} strategy).
For scalability purposes, the authors also suggest an alternative query strategy which does not use the VOI criterion to rank a set of candidate comparisons. At each preference elicitation just one query is considered, namely the comparison between the configuration $\x^*$ with greatest expected utility and the solution $\x^{EL}$ maximizing the expected loss of recommending $\x^*$ instead of $\x^{EL}$ (\emph{simplified VOI} strategy). 

Unlike CLEO, GSM is conceived for instances characterized by purely
discrete attributes, and cannot tackle preference elicitation tasks
over hybrid domains. In our experiments (Sec.~\ref{sec:exp}), an
empirical comparison of CLEO w.r.t. GSM is thus performed over a
simplified experimental setting involving discrete decisional
attributes only.


\subsection{Constraint-based preference elicitation}


The work in~\cite{softConstr10} articulates the user preferences in terms of \textit{soft} constraints and introduces constraint optimization problems where the DM preferences are not completely known before the
solution process starts. In soft constraints 
each assignment to the variables of one constraint is associated with a
preference value taken from a preference set. The preference value
represents the level of desirability of the assignment to the
variables of the constraint. As the preference score is associated to
a partial assignment to the problem variables, it represents a
\emph{local} preference value.  The desirability of a complete
assignment is defined by a \emph{global} preference score, computed by
applying a combination operator to the local preference values. A set
of soft constraints generates an order (partial or total) over the
complete assignments of the variables of the problem. Given two
solutions of the problem, the preferred one is selected by computing
their global preference levels.
Preference elicitation strategies have been 
introduced~\cite{softConstr10} 
to deal with scenarios where preference information is partially unknown. 
Some of the local preference values attached to soft
constraints are assumed to be missing, and the DM is asked for an
explicit feedback on specific assignments for these constraints, in
terms of score values quantifying her preference for a certain
assignment. 
In comparison to this approach based on the Constraint Programming formalism, 
CLEO assumes a much more limited amount of initial knowledge about the problem at hand. 
In~\cite{softConstr10}, decision variables, soft constraint topology and structure are assumed to be known in
advance and the incomplete initial information consists of missing local preference values only. 
CLEO assumes complete ignorance about the structure of the constraints over the decisional variables of the user. The initial problem knowledge is limited to a set of catalog attributes. CLEO extracts
the decisional items of the DM from the set of catalog attributes and learns
the weighted constraints constructed from them modeling the DM preferences. 

Furthermore, the technique in~\cite{softConstr10} is based on local elicitation queries,
with the final user asked to reveal her preferences about assignments for
specific soft constraints. Global preferences or bounds for global preferences
associated to complete solutions of the problem are derived from the local
preference information. CLEO goes in the opposite direction: it asks
the user to compare complete solutions and learns local utilities (i.e., the
weights of the soft constraints of the logic formula) from global preference values. In
many cases, recognizing appealing or unsatisfactory global solutions may be
much easier than defining local utility functions, associated to partial solutions. For example, while scheduling a set of activities, the evaluation of
complete schedules may be more affordable than assessing how specific ordering choices between couples of activities contribute to the global preference
value. Furthermore the algorithm in~\cite{softConstr10} asks the DM 
for quantitative evaluations of partial solutions: she does not just rank couples of activities, she provides score values quantifying her preference for the
partial activity rankings, a much more demanding task. Finally, the approach in~\cite{softConstr10} assumes consistent and accurate quantitative feedback from the DM. Under this assumption, the optimality of the recommended solution is guaranteed. However, this approach cannot be applied in our realistic experimental setting characterized by the noisy human feedback. 


\section{Experimental results}
\label{sec:exp}
  
The following empirical evaluation demonstrates that CLEO can handle realistic preference 
elicitation tasks defined over hybrid domains and with uncertain human feedback.
No alternative algorithm capable of tackling these preference elicitation tasks is currently available (see Sec.~\ref{sec:related}). To overcome this limitation, our experimental work consists of two phases.
First, CLEO is tested over a couple of realistic preference elicitation tasks with the above features. 
For this purpose, a benchmark of Max-SMT problems is defined, involving both discrete and continuous decisional variables. In a second step, a set of simplified synthetic problems with discrete decisional variables only is introduced, in order to compare CLEO with the existing preference elicitation algorithms. In particular, we consider Boolean decisional attributes only and generate a set of synthetic Maximum-Satisfiability (Max-SAT) benchmarks. 
In this simplified setting, the benchmarking preference elicitation algorithm is the method by Guo and Sanner~\cite{GSMalgorithm10}. 

For the experiments performed, the mapping function $\lmtPSI$ in CLEO projects configurations into the space of all possible conjunctions of up to three atomic constraints (i.e., $d=3$). 
The next section describes
the well-known noisy response model used in both Max-SMT and Max-SAT
experiments for simulating inaccurate and inconsistent feedback
provided by the DM during the preference elicitation process.


\subsection{Noisy response model for human feedback}

In the experiments the feedback from the user is assumed to be affected by the inaccuracies and inconsistencies.  
The user ranks configurations $\lmta$ based on a latent utility function $f(\lmta)$.
In particular, configuration $\lmta^i$ is preferred to configuration $\lmta^j$, i.e., $\lmta^i \succ \lmta^j$, if and only if $f(\lmta^i) > f(\lmta^j)$. However, each evaluation $f_i = f(\lmta^i)$ is corrupted by additive independent and identically distributed (IID) Gaussian noise $\varepsilon_i \sim \mathcal{N} (0,\sigma^2_{noise})$, resulting in a noisy utility value $y_i = f_i +\varepsilon_i$.  

Under the assumption of independent and identically distributed Gaussian noise, the probability that the user prefers configuration $\lmta^i$ to configuration $\lmta^j$ is defined as follows:
\begin{align}
	\label{eq:probit_pairwise_comp}
	 & \mathrm {P} (\lmta^i \succ \lmta^j | f_i, f_j) = \mathrm {P} (y_i > y_j | f_i, f_j)  =   \nonumber \\
         & \mathrm {P} (f_i +\varepsilon_i > f_j +\varepsilon_j)  =  \mathrm {P} (\varepsilon_i - \varepsilon_j > f_j - f_i) 
\end{align}
The quantity $\delta = \varepsilon_i - \varepsilon_j$ is the difference of two IID Gaussian variables with zero-mean and variance $\sigma^2_{noise}$, and therefore follows the Gaussian distribution $\mathcal{N} (0, 2\sigma^2_{noise})$.
By computing the standardized variable $z= \delta / (\sqrt{2} \sigma_{noise})$, Eq.~(\ref{eq:probit_pairwise_comp}) can be rewritten as: 
$$
\mathrm {P} (\varepsilon_i - \varepsilon_j > f_j - f_i) = 1 - \Phi\left( \frac{f_j - f_i}{\sqrt{2} \sigma_{noise}} \right)
$$
where $\Phi$ is the cumulative distribution function of the standard normal distribution. 

The above user response model, linking pairwise comparisons to a continuous latent utility function, has been widely used in the economic and psychological studies to describe the individual choice behaviour of humans~\cite{rank_algo_11, pairwiseComp11,economicchoices01}. It is known as the Thurstone-Mosteller or Probit model. In our experimental setting $\sigma^2_{noise}$ is fixed to $10$, to have noise values comparable with the latent utility values $f(\lmta)$.


\subsection{Realistic preference elicitation tasks over hybrid domains}

CLEO is tested over a benchmark of Max-SMT problems, formulating realistic preference elicitation tasks.
The Max-SMT tool used for the experiments is the ``Yices" solver~\cite{yices_paper} (version 1.0), which is publicly available at \url{http://yices.csl.sri.com/} (as of August 2015).
Each point of the curves depicting our results is the median value over 400 runs with different random seeds.

Max-SMT is a recent research area. Even if existing
results~\cite{NieOli06} indicate that Max-SMT solvers can efficiently
address real-world problems, to the best of our knowledge 
no well-established publicly available Max-SMT benchmarks exist and 
preference elicitation tasks have not been encoded into Max-SMT instances yet.

In this work, we modelled a \emph{scheduling} problem as a Max-SMT
instance, where the DM expresses her preferences about the candidate
schedules of a set of jobs. In the spirit of real-world 
recommendation tasks, we also design a \emph{housing} problem aimed
at selecting a location for building a house. The formulation consists
of both unknown soft constraints representing the user preferences and
known hard constraints defining the feasible search space. The housing problem is challenging, 
due to complex non-linear relationships among decision variables. For example, the variable encoding the cost of the 
location is defined as a function of the remaining decision variables.
The results obtained by CLEO over both the preference elicitation tasks are discussed below.


\subsubsection{Scheduling problem}

A set of five jobs must be scheduled over a given period of
time. Each job has a fixed known duration, the atomic constraints define the
overlap of two jobs or their non-concurrent execution. The user unknown 
utility function is generated by selecting uniformly at random weighted conjunctions of atomic constraints. The solution of the problem is a schedule
assigning a starting date to each job and maximizing the utility, where
the utility of the schedule is the sum of the weights of the satisfied 
constraints of the user utility function.  
The atomic soft constraints define temporal constraints 
by using the difference arithmetic theory. In detail, let
$s_i$ and $d_i$, with $i=1\ldots5$, be the starting date and the
duration of the i-\textit{th} job, respectively. If $s_i$ is scheduled
before $s_j$, the constraint expressing the overlap of the two jobs is
$ s_j - s_i < d_i$, while their non-concurrent execution is encoded by
$ s_j - s_i \geq d_i$. Let us note that there are 40 possible constraints for
a set of 5 jobs. The maximum size of the soft constraints 
is assumed to be three. The weights of soft constraints are distributed uniformly at random in the range $[1, 100]$. 

CLEO is tested over a benchmark of randomly generated utility functions according to the
couple (\emph{number of decisional features, number of soft constraints}). The decisional features are the atomic constraints appearing in the soft constraints. We generate functions for the following values: $\{(5,3), (10,6), (15,9)\}$. Each DM utility has at least two
soft constraints with a size of three. Let's underline once more that utility functions with more that few factors or factors with many terms are unrealistic when considering human DM~\cite{mil56}.

\begin{figure}
  	\centering
  	\centerline{
		\includegraphics[width=.33\textwidth]{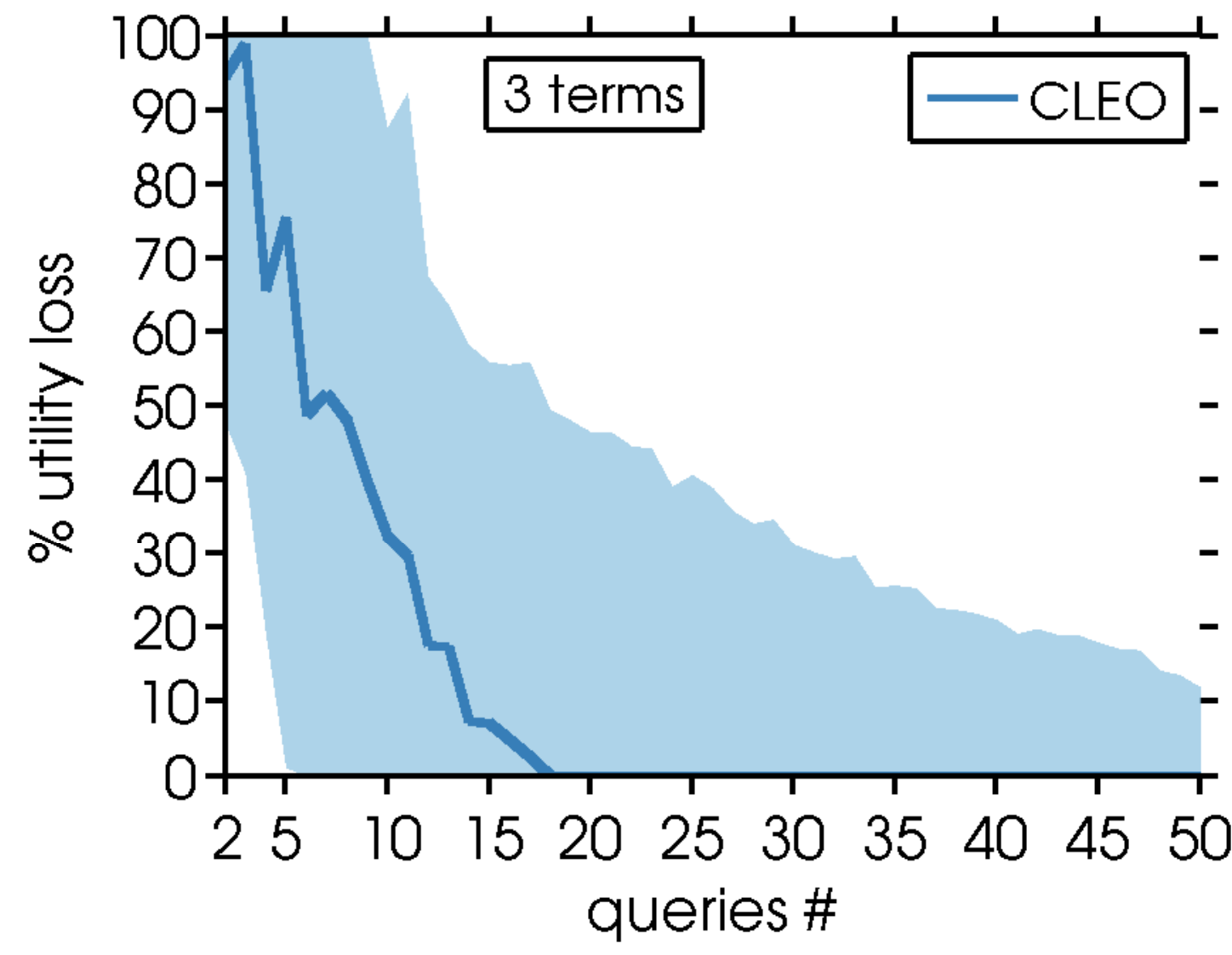}
		\includegraphics[width=.33\textwidth]{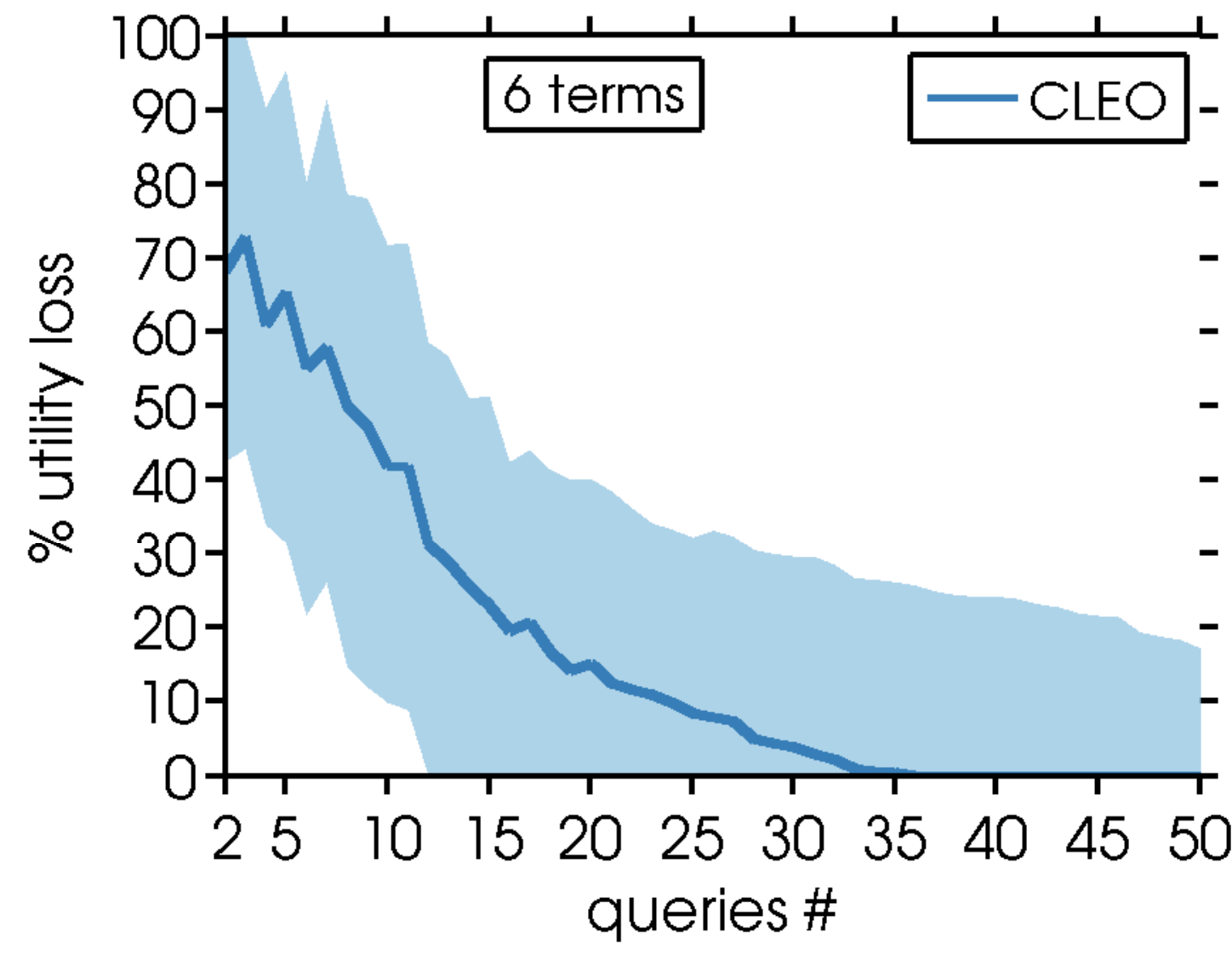}
		\includegraphics[width=.33\textwidth]{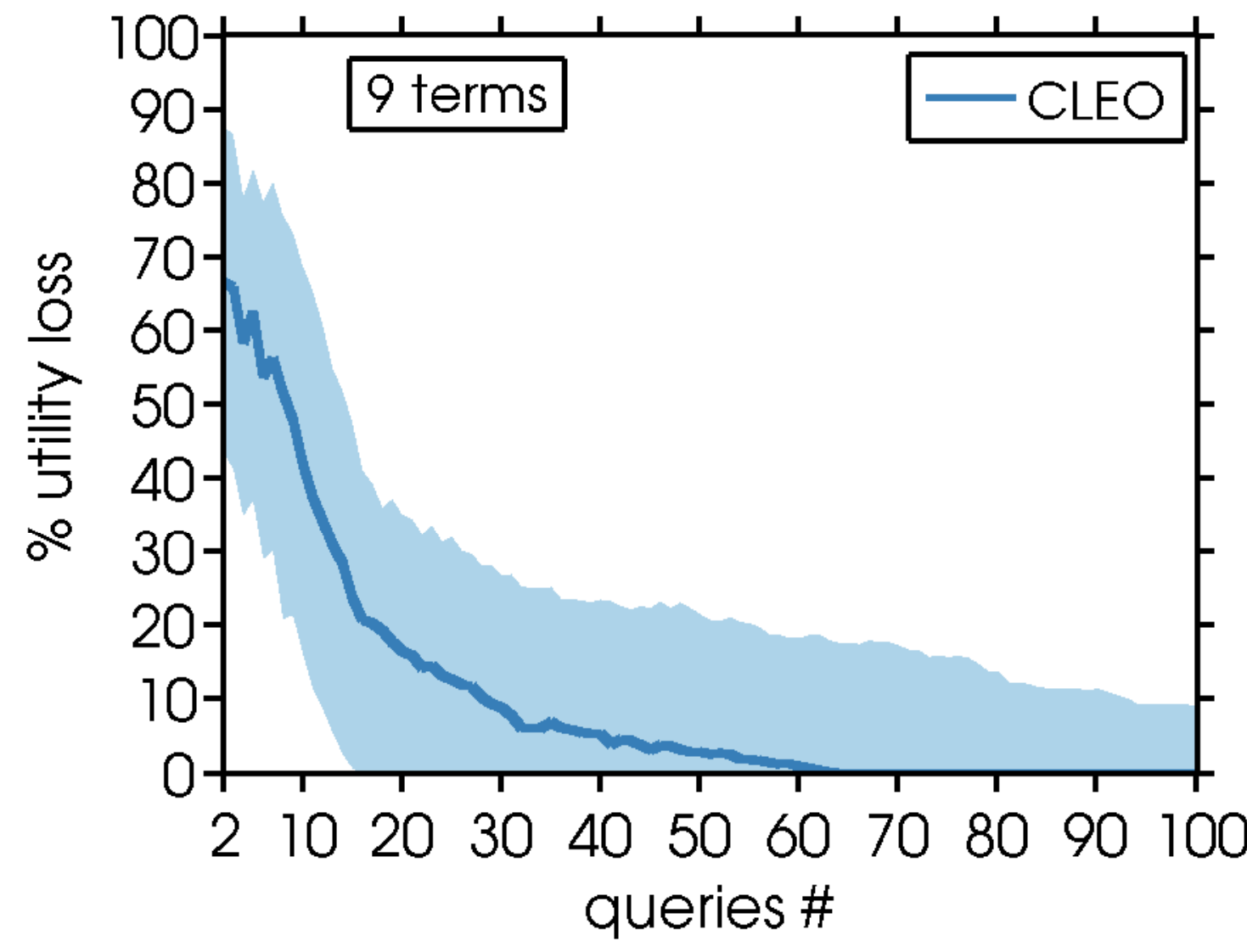}
    		   }
	\caption{Performance of CLEO in solving the scheduling problem. The \textit{y}-axis reports the percentage utility loss, while the \textit{x}-axis contains the number of pairwise-comparison queries asked so far. The curve reports the median values observed over 400 runs of CLEO, while the shaded area denotes the range among the 25th and the 75th percentiles. Please note the different range of the \textit{x}-axis in the case of nine soft constraints.}
	\label{fig:scheduling_perf}
\end{figure}

Results of the experiments are shown in Figure~\ref{fig:scheduling_perf}. The
\textit{y}-axis reports the percentage utility loss measured in terms
of deviation from the utility of the DM preferred solution, while the
\textit{x}-axis contains the number of pairwise comparisons asked so
far. The curves report the median values observed over 400 runs, while
the shaded area depicts the interquartile range measuring the
dispersion around the median.

As expected, the learning problem becomes more challenging for an
increasing number of soft constraints. However, results are promising,
as a substantial improvement in the quality of the recommended
solution is achieved by CLEO when additional queries are asked to the
DM (anytime property). Furthermore, CLEO identifies the DM preferred
solution in all cases. In detail, with the realistic cases of
three and five soft constraints, less than 35 pairwise comparisons are
asked to the DM to identify her preferred solution.  With 9 soft
constraints, $64$ pairwise comparisons are required on average to
recommend the DM preferred solution. However, with 40 queries, a
percentage utility loss within $5.5\%$ is obtained.  The shaded area
shows that CLEO identifies the DM preferred solution quite
consistently when increasing the number of queries (the interquartile
range is within $25\%$ after $35$ queries even in the case of nine soft constraints).

\subsubsection{Housing problem}

We consider a customer planning to build her own house and judging potential housing locations provided by a real estate company (henceforth the \emph{housing} problem).
There are different locations available where the customer may potentially build her house. The locations are characterized by different housing values, prices, constraints about the design of the building (e.g., usually in the city center you cannot have a family house with a huge garden and pool), etc.
The customer may formulate her judgments by considering a description 
of the housing locations based on a predefined set of parameters, including, e.g., crime rate, distance from downtown, location-based taxes and fees, public transit service quality, walking and cycling facilities, proximity to commercial facilities or green areas, etc. 
Many of these parameters may be uninformative, as they do not represent any decisional criterion for the customer. 
Furthermore, hard constraints defining the feasible locations may be specified in advance, e.g., cost bounds stated by the user or building design requirements asserted by the company. 

In our experiments, the formulation of the housing problem is as follows. 
The set of catalog attributes is listed in Table~\ref{tab:housingfFeat}. 
\begin{table}[ht]
	\caption{Catalog attributes for the housing problem.} 
	\begin{center}

	\begin{tabular}{r l l} 
	\hline\hline 
	num  & attribute & type \\ [0.5ex] 
	\hline 
	1    & house type          &  		  	   ordinal  \\ 
	2    & garden              &    		   Boolean        \\         
	3    & garage		   & 		  	   Boolean  \\
	4    & commercial facilities in the neighborhood & Boolean \\
	5    & public green areas in the neighborhood    & Boolean \\ 
	6    & cycling and walking facilities in the neighborhood   & Boolean   \\
	7    & distance from downtown      	  & 	   numerical \\
	8    & crime rate		          &	   numerical \\
	9   & location-based taxes and fees      &    	   numerical \\
	10   & public transit service quality index    &   numerical \\
	11   & distance from high schools 	  &   	   numerical \\
	12   & distance from nearest free parking     &	   numerical \\
	13   & distance from working place            &    numerical \\
	14   & distance from parents house            &    numerical \\ \hline
	15   & price                              & 	   numerical \\  [1ex] 

	\hline 
	\end{tabular}
	\end{center}
	\label{tab:housingfFeat} 
\end{table}
A set of ten hard constraints (Table~\ref{tab:housingHC}) defining feasible housing locations and  
 known in advance is considered. The hard constraints are stated by the customer (e.g., cost bounds) or by the company (e.g, constraints about the distance of the available locations from  
 user-defined points of interest). Let us note that constraints 5, 6, 7 define a linear bi-objective problem among distances from user-defined points of interest.
Prices of potential housing locations are defined as a function of the other attributes. For example, price increases if a semi-detached house rather than a flat is selected or in the case of green areas in the neighborhood. On the other side, e.g., when crime index of potential locations increases, price decreases.
Soft constraints are represented by weighted conjunctions of both predicates in the linear arithmetic theory and Boolean variables, in the case of attributes number $2, 3,\dots, 6$ in Table~\ref{tab:housingfFeat}. For example, one predicate may model the preference for a location with distance from nearest free parking smaller than a given threshold, while a Boolean variable encodes, e.g., the aspiration for houses with garage.

\begin{table}[ht]
	\caption{Hard feasibility constraints for the housing problem. Parameters \textit{$\rho_i$}, $i = 1 \dots 13$, are threshold values specified by the user or by the sales personnel, depending on who states the hard constraint which they refer to. } 
	\begin{center}
	\begin{tabular}{r l l} 
	
	\hline\hline 
	num  & hard constraint       \\ [0.5ex] 
	\hline 
	1    & price $\leq$ \textit{$\rho_1$}   	  \\        
	2    & location-based taxes and fees $\leq$ \textit{$\rho_2$} $=>$ \textit{not} public 
             green ares in the  \\
             & neighborhood \textit{and not} public transit service quality index $\leq$  \textit{$\rho_3$} \\   
	3    &  commercial facilities in the neighborhood $=>$ \textit{not} 
             (garden \textit{and} \\
		 &  garage) \\
	4    &  crime rate $\leq$ \textit{$\rho_4$} $=>$ distance from downtown $\geq$ \textit{$\rho_5$} \\
	5    &  distance from working place + distance from parents house $\geq$ \textit{$\rho_6$} \\ 
	6    &  distance from working place + distance from high schools $\geq$ \textit{$\rho_7$} \\ 
	7    &  distance from parents house + distance from high schools $\geq$ \textit{$\rho_8$} \\ 
	8    &  distance from nearest free parking $\leq$ \textit{$\rho_9$} $=>$ \textit{not} public
 	     green areas  \\ 
             & in the neighborhood \\
	9    &  distance from parents house $\leq$ \textit{$\rho_{10}$}   $=>$ distance from downtown $\geq$ \\ 
         &  \textit{$\rho_{11}$} \textit{and} crime rate $\geq$ \textit{$\rho_{12}$} \\
	10   & garden $=>$ house type $\geq$ \textit{$\rho_{13}$} \\ [1ex] 

	\hline 

	\end{tabular}
	\end{center}
	\label{tab:housingHC} 
\end{table}


We generated a set of 40 predicates, i.e., atomic constraints. The 
user unknown utility function is composed of soft constraints with two or
three predicates, with at least one soft constraint with three
predicates. The maximum number of predicates in a soft constraint is
assumed to be known. The weights of soft constraints are integer
values selected uniformly at random in the range $[1, 100]$.

\begin{figure}
  \centering
  \centerline{
    \includegraphics[width=.33\textwidth]{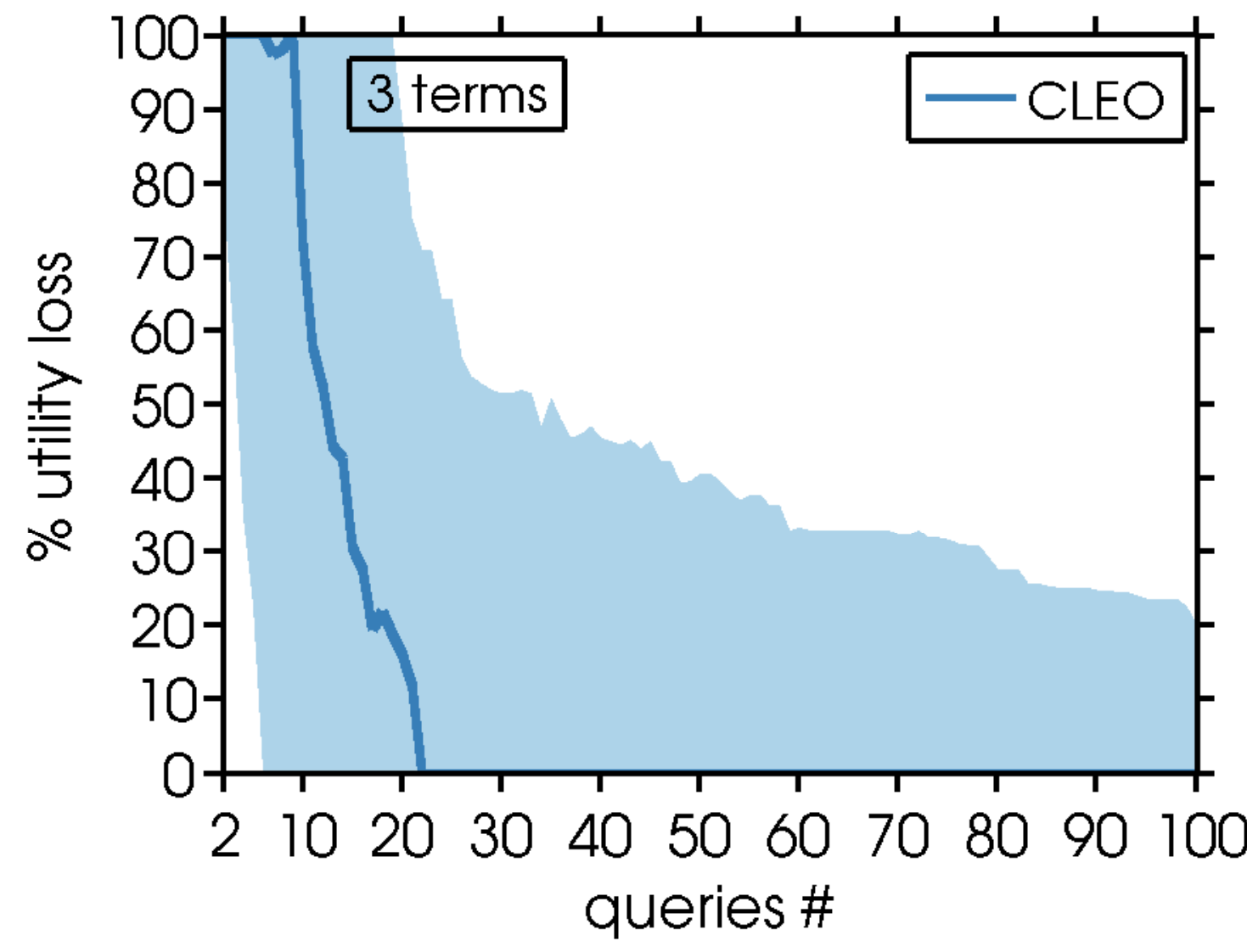}
    \includegraphics[width=.33\textwidth]{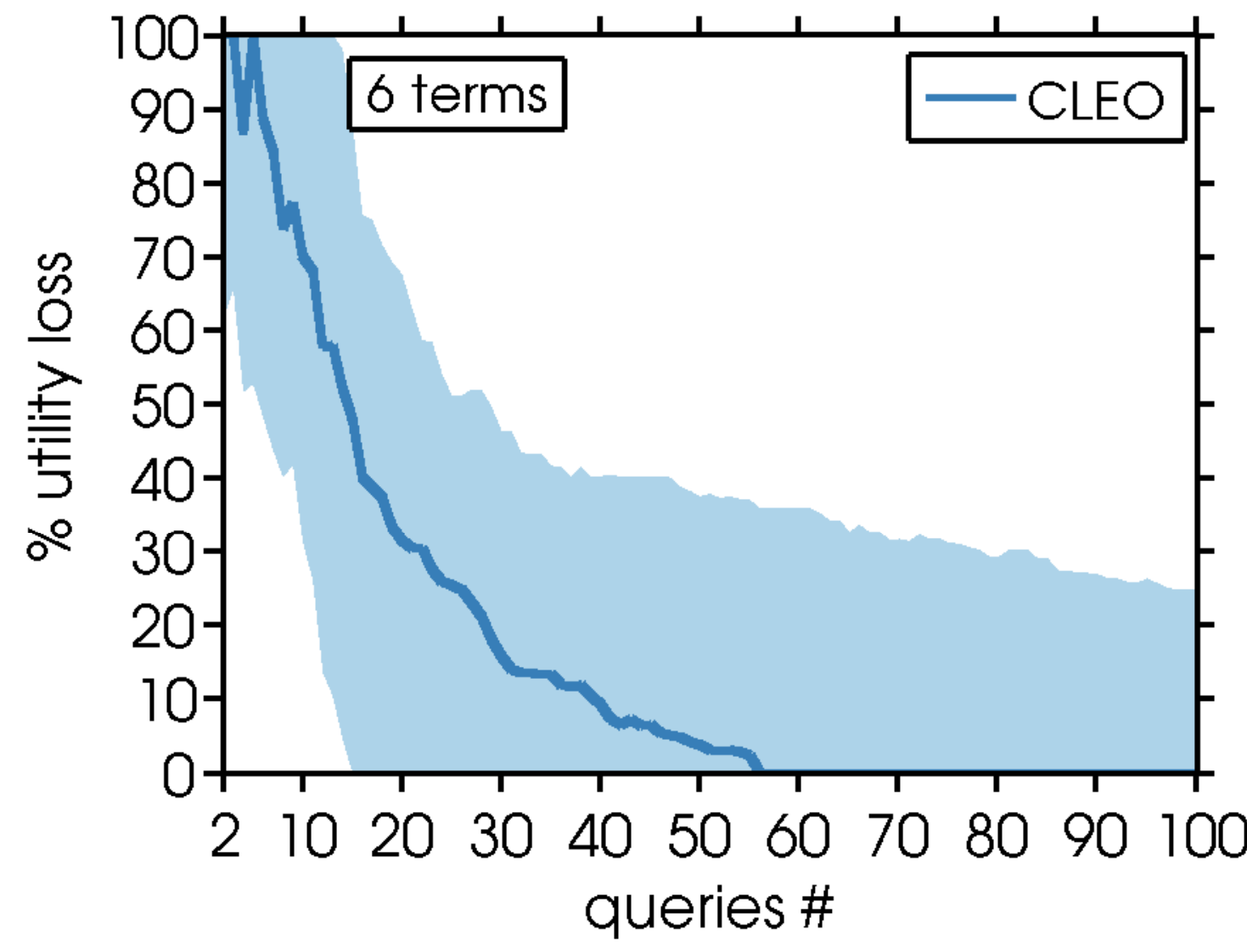}
    \includegraphics[width=.33\textwidth]{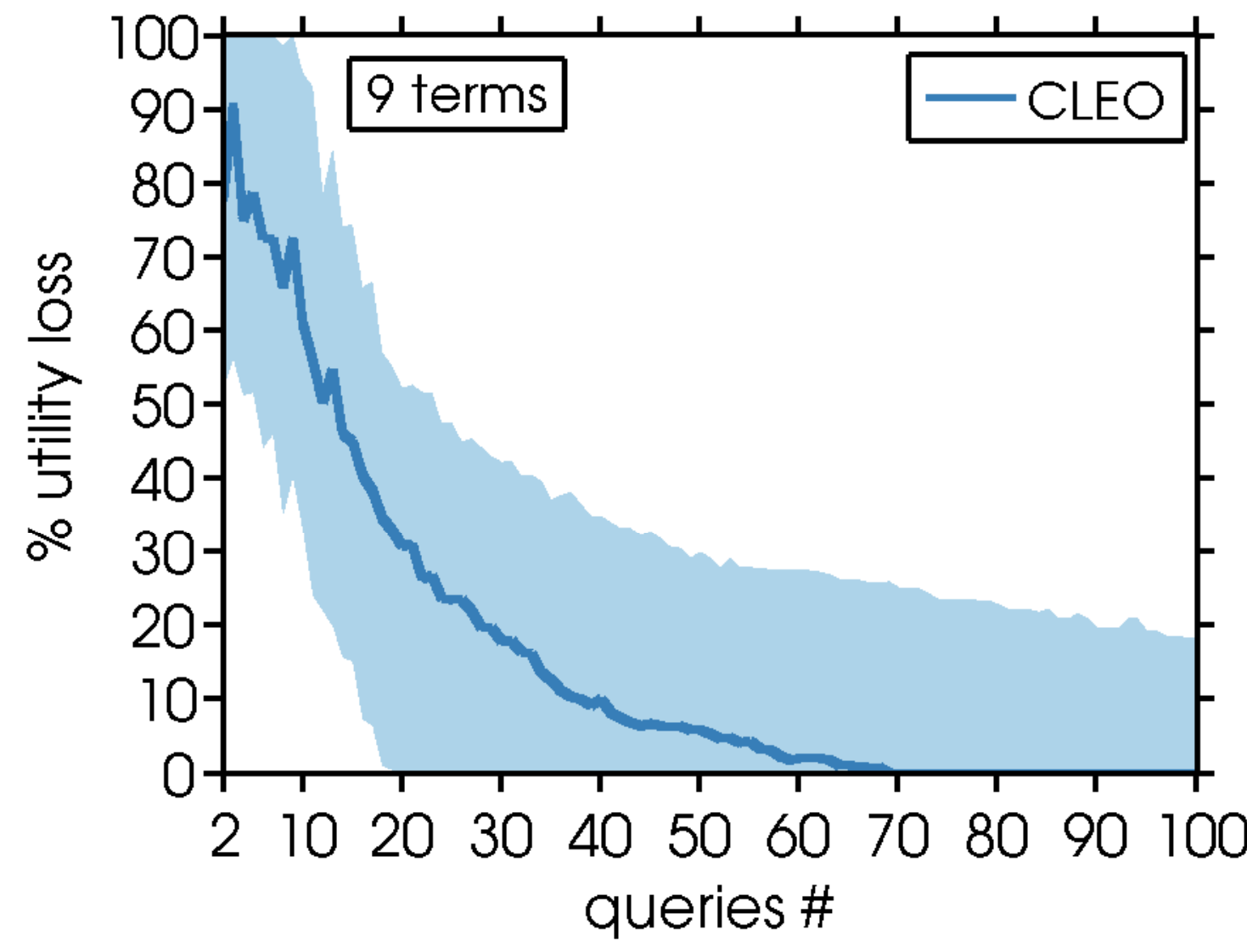}
    }
  \caption{Performance of CLEO while solving the housing problem.} 
  \label{fig:Housing_perf}
\end{figure}

Fig.~\ref{fig:Housing_perf} reports the results over a benchmark of
400 randomly generated utility functions for each of the following
instantiations of the couple (\emph{number of decisional features,
  number of soft constraints}): $\{(5,3), (10,6), (15,9)\}$, where the
decisional features are the predicates appearing in the soft
constraints. The promising results observed for the scheduling problem
are confirmed, even though the housing problem is much harder, due to
complex non-linear interactions among the decisional attributes. When
increasing the number of queries asked, the quality of the solution
rapidly improves and CLEO identifies the DM preferred configuration in
all the cases.  On average, 22 and 69 queries are needed by CLEO to
converge to the DM preferred solution in the case of three and nine
soft constraints, respectively. Let us note again that utility
functions involving nine soft constraints are quite unrealistic and
are considered here just for testing the scalability of CLEO.

The dispersion of the performance values keeps decreasing when increasing the number of queries asked, showing that CLEO recommends better quality solution more consistently.
However, in the case of three soft constraints, the interquartile range observed when CLEO converges is equal to $70.8\%$. With 40 queries, the dispersion decreases down to $45.4\%$.
These values are rather large. 
A deeper investigation of CLEO results revealed that the observed data dispersion is heavily affected by some runs where the solution quality does not improve when asking additional feedback to the DM. In these runs CLEO cannot generate queries informative enough to recover from suboptimal initial choices. Smarter queries strategies could be studied in order to tackle these cases, as discussed in Sec.~\ref{sec:disc}. 


\subsection{Experimental comparison with the state-of-the-art}
\label{sec:soa_comparisons}

Since existing methods cannot handle the preference elicitation tasks over hybrid domains defined in the previous section, for a comparison with the state-of-the-art we focus on Boolean attributes only. 
With this choice, the atomic constraints are just the Boolean attributes, and more complex soft constraints expressing the DM preferences are Boolean terms in plain propositional logic. That is, each soft constraint is the conjunction of (up to three) Boolean attributes and the unknown DM utility function is a \emph{weighted Maximum Satisfiability} (Max-SAT) instance consisting of the weighted combination of the Boolean terms. The benchmarking algorithm is the GSM method~\cite{GSMalgorithm10} described in Sec.~\ref{subsec:Bayesian}.

A benchmark of random utility functions is generated for (\emph{number
  of Boolean attributes, number of terms}) equal to
$\{(5,3), (10,6), (15,9)\}$. Each utility function has two constraints
with maximum size (three). Constraint weights are integers selected
uniformly at random in the interval $[-100, 0) \cup (0, 100]$. 

All the query selection strategies suggested in~\cite{GSMalgorithm10}
for the GSM method have been tested in our experimental setting. For
each of the three test cases $\{(5,3), (10,6), (15,9)\}$, we report
here the results of the query strategy with best performance. However,
with more than five attributes, the most sophisticated Bayesian query
strategies proposed in~\cite{GSMalgorithm10} are too slow, as pointed
out also by the authors themselves and empirically verified in our
preliminary experiments. They have thus been included in the
$(5,3)$ case only. Based on our results, the best query strategy
are the ``restricted informed value of information (VOI)" for the test
case $(5,3)$ and the ``simplified VOI" for both remaining test cases.

\begin{figure}
  \centering
  \centerline{
    \includegraphics[width=.333\textwidth]{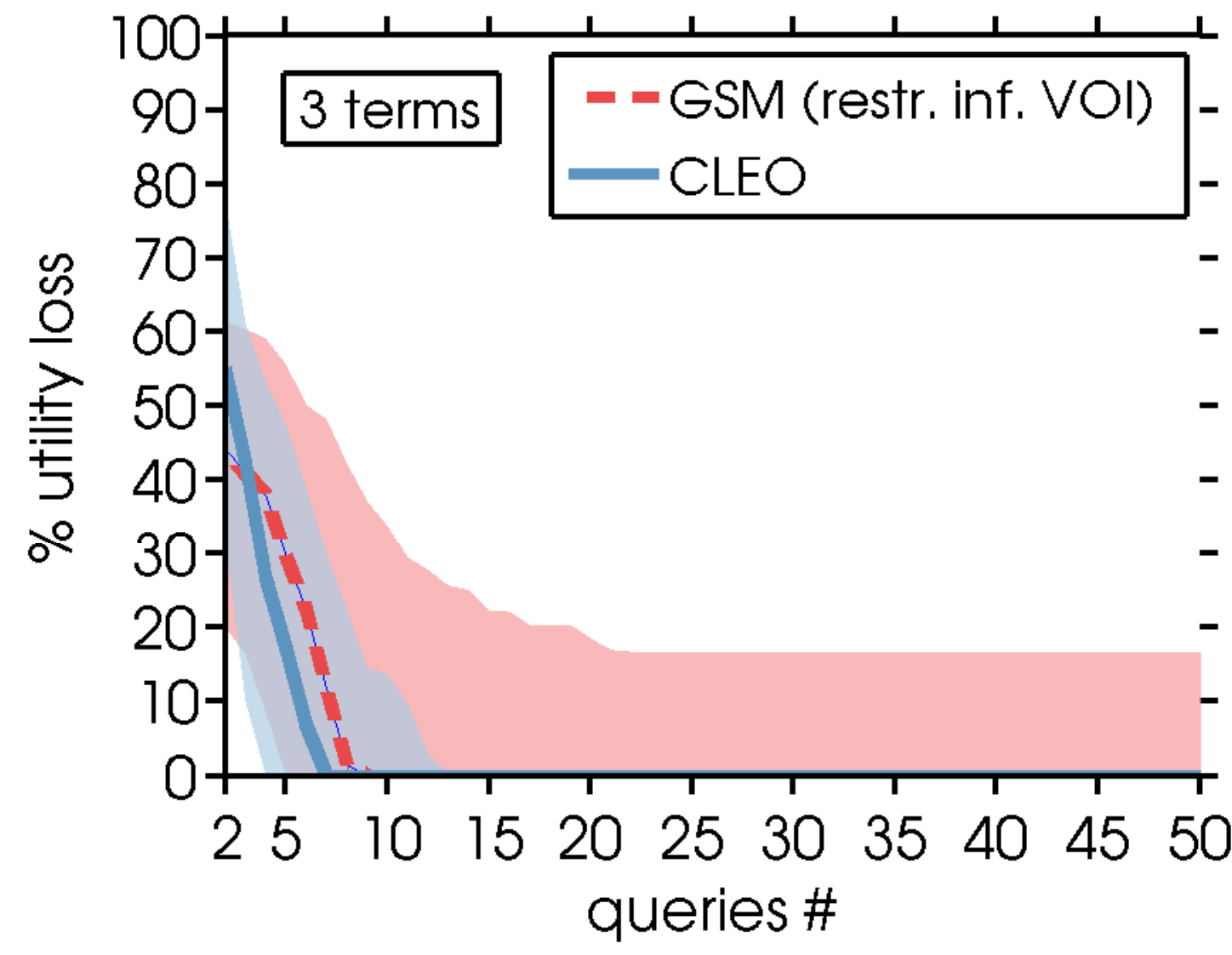}
    \includegraphics[width=.333\textwidth]{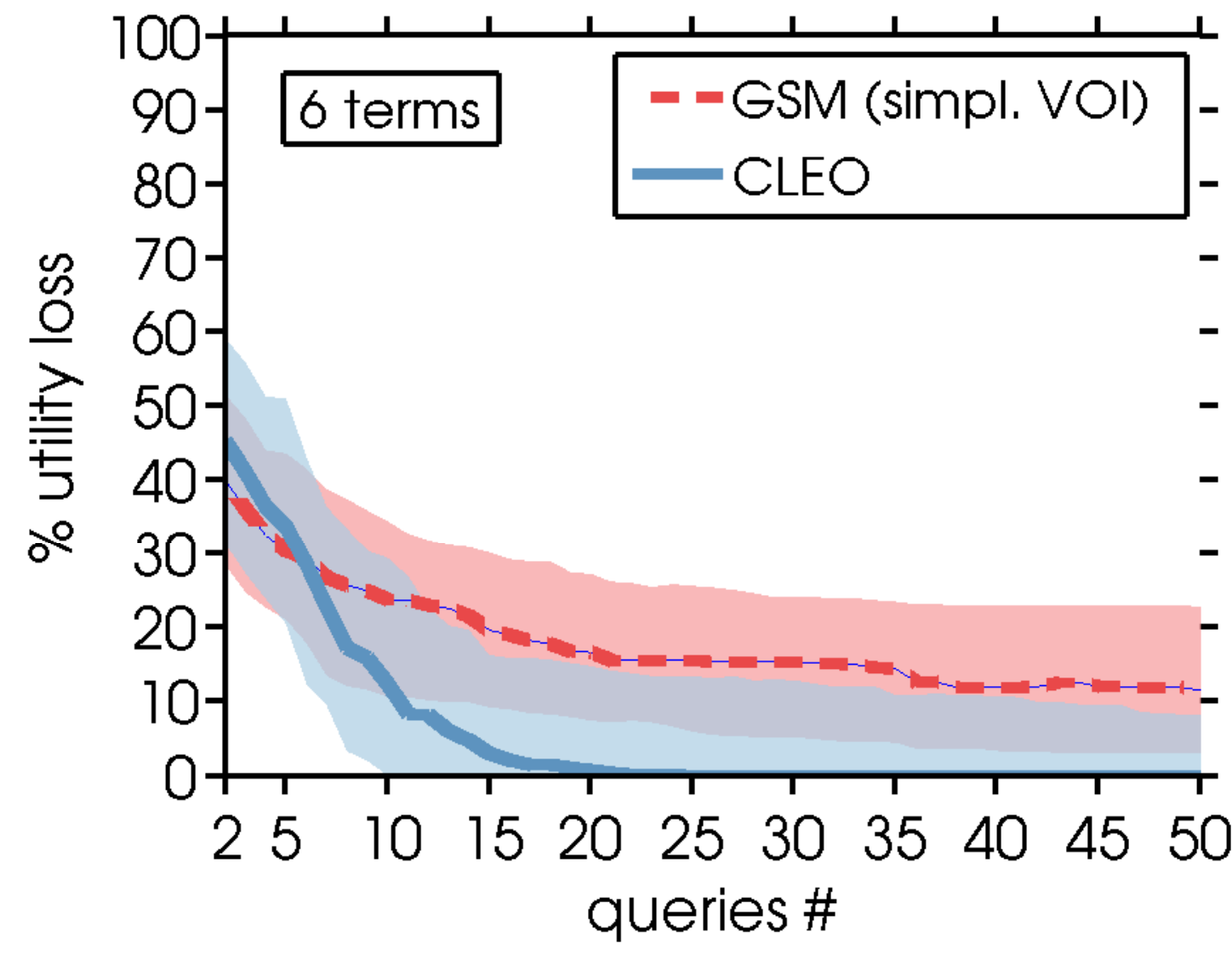}
    \includegraphics[width=.333\textwidth]{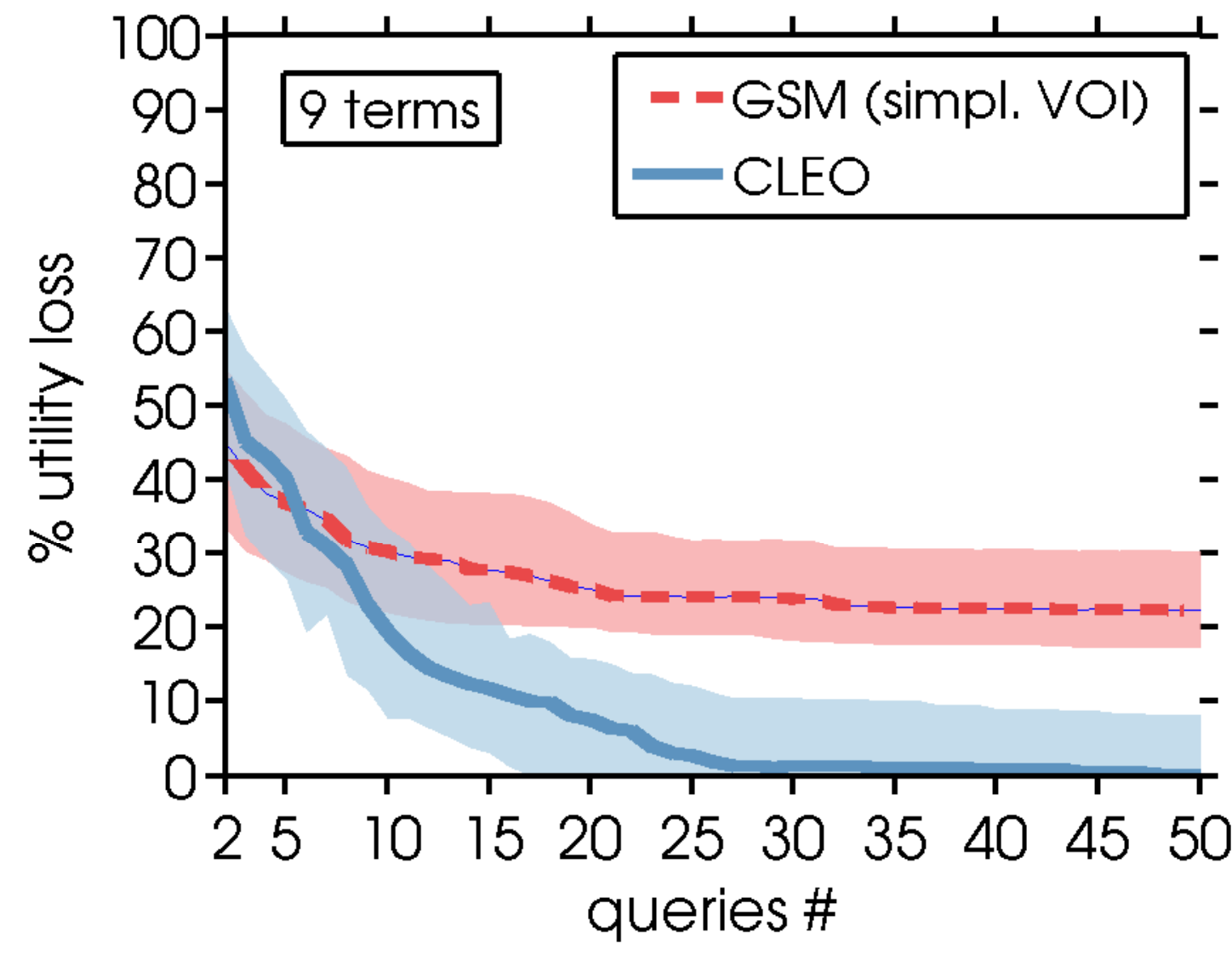}
    }
  \caption{Performance of the CLEO (darker solid line or blue solid line if viewed in colour) and GSM (lighter dashed line or red dashed line if viewed in colour) algorithms over the Boolean problems. The \textit{y}-axis reports the percentage utility loss, while the \textit{x}-axis contains the number of pairwise comparisons asked so far. The curves report the median values observed over 200 runs, while the shaded area denotes the range among the 25th and the 75th percentiles of the observations. Best viewed in colour.}
  \label{fig:Boolean_perf}
\end{figure}

Fig.~\ref{fig:Boolean_perf} reports the percentage utility loss of the recommended configuration w.r.t the DM preferred solution for an increasing number of pairwise comparisons asked so far. 
The curves report the median values observed over 200 runs for CLEO (darker solid line or blue solid line if viewed in colour) and GSM (lighter dashed line or red dashed line if viewed in colour). The shaded areas depict the interquartile range measuring the dispersion around the median. 

The search space of the simplest problem with five Boolean attributes contains just 32 candidate configurations, thus any strategy asking more than few questions is not competitive with na\"{i}ve exhaustive search. 
On average, seven and nine queries are asked to the DM by CLEO and GSM for discovering her preferred solution. 
However, with 12 (or less) queries, the CLEO and GSM performance are statistically equivalent under
a Two-sided Wilcoxon signed-rank test with a Bonferroni-corrected significance level of $10^{-3}$. 
With more than 12 queries, there is statistical evidence for better results by CLEO, due to the much more unstable behavior of the GSM method: after 14 queries CLEO consistently identifies the DM preferred solution with a null interquartile range (IQR), while the IQR of the GSM results remains above 16.6\%.

The more challenging test cases are represented by the problems with 10 and 15 Boolean attributes, where the search space size is 1024 and 32768, respectively, preventing the application of exhaustive search techniques. 
In both these cases, the performance of CLEO is much better than that of GSM.

In detail, with 10 Boolean attributes, CLEO on average asks 25
pairwise comparisons to the DM for identifying her favourite solution,
while the average percentage utility loss of the configuration
recommended by GSM remains above 10\% even if 50 queries are asked to
the DM. With 16 queries, the CLEO curve is within 2\%, against a value
of around 19\% observed for GSM.  The performance difference between
CLEO and GSM is significant at $10^{-3}$ level after eight queries,
and the significant level goes to $10^{-11}$ after 15 queries.

An analogous situation is observed for the $(15,9)$ test case. The
solution returned by CLEO has an average loss of less than 2\% after
26 queries and less than 1\% after 38 ones. On the other hand, after
50 queries, GSM recommends on average solutions with a loss still
above 22.3\%. The performance difference after the first seven queries
is statistically significant with a $10^{-3}$ level, which goes to
$10^{-10}$ after ten queries. 


\section{Conclusions}
\label{sec:disc}

This paper introduces CLEO, a preference elicitation algorithm that, unlike existing 
approaches, handles preference elicitation tasks defined over hybrid domains and with uncertain human feedback.
A combinatorial formulation of the unknown DM utility function is adopted. CLEO consists of an incremental procedure, iteratively optimizing the learned approximation of DM utility function
to generate candidate solutions 
and refining the approximation based on the human feedback received.
Simple pairwise comparison queries are asked to the DM. 

CLEO assumes \emph{very limited initial knowledge}. 
In detail, since different decision 
makers usually have different decisional criteria, the algorithm just assumes a
set of \emph{catalog} attributes describing the candidate
configurations. The DM preferences are expressed by soft constraints
over the attributes values.  However, only a small subset of catalog
attributes (and, by consequence, of soft constraints defined on them)
may be relevant for a specific DM, resulting in a sparse learning
setting, both in the number of relevant attributes and soft
constraints. The algorithm employs 1-norm regularization, which enforces
sparsity of the learned function, in order to identify the relevant
attributes and constraints.

The learned function is a set of weighted soft constraints involving both discrete and continuous-valued attributes. 
The configuration maximizing the weights of the satisfied constraints is recommended to the DM. To identify this configuration, a Max-SMT solver is used. 
CLEO is a generic framework, enabling the adoption of well-assessed learning methods and Max-SMT solvers.

Experimental results on realistic preference elicitation tasks
demonstrate the effectiveness of CLEO in focusing towards the optimal
solutions, its robustness, as well as its ability to recover from
suboptimal initial choices.  Our experiments involve preference
elicitation tasks over hybrid domains, with uncertain human feedback,
(known) hard constraints limiting the set of feasible configurations
and complex non-linear interactions among the decisional attributes
(e.g., the cost attribute in the case of the housing problem). CLEO
has also been compared with a state-of-the-art Bayesian preference
elicitation approach in a simplified setting with purely discrete
attributes. The experimental results show that CLEO outperforms the
benchmarking algorithm, with the performance difference becoming more
pronounced when increasing the complexity of the preference
elicitation task.

CLEO can be generalized in a number of directions. The learning stage employs a 
ranking loss function based on pairwise preference evaluation. More complex ranking losses
have been proposed in the literature (see for
instance~\cite{ChaetAl08}), especially to increase the importance of
correctly ranking the highest scoring solutions, and could be combined with
1-norm regularization.

\emph{Active learning} is a hot research area and a broad range of
different approaches has been proposed (see~\cite{Set09} for a
review). The simplest and most common framework is that of
~\emph{uncertainty sampling}: the learner queries the instances on
which it is least certain. However, the ultimate goal of a
recommendation or optimization system is selecting the best
instance(s) rather than correctly modeling the underlying utility
function. The query strategy should thus tend to suggest good
candidate solutions and still learn as much as possible from the
feedback received. Typical areas where research on this issue is quite
popular are single- and multi-objective interactive
optimization~\cite{MooInt08} and information
retrieval~\cite{RadJoa07}.  The need to trade-off multiple
requirements in this active learning setting is addressed
in~\cite{DiversityAndDensity07} where the authors consider relevance,
diversity and density in selecting candidates.  Our future research
will consider the application of these active learning techniques.
The performance of our method indeed depends on the trade-off between
the identification of candidates solutions satisfying the DM (i.e.,
solutions optimizing the current learned preference model) and the
generation of \emph{informative} training examples for the following
refinement of the learned model.

In the context of preference elicitation, Bayesian approaches are
attractive as they quantify the uncertainty in the learned DM utility
models and provide a principled approach to estimate the value of the
information obtained by asking a certain query to the DM. In
particular, the value of the information estimates the extent to which
a certain query helps in improving the quality of the learned
preference model.
The value of information is exploited to design efficient query
strategies consisting of informative queries, see, e.g., the
GSM~\cite{GSMalgorithm10} algorithm we use as benchmark in the
experimental comparisons. Adapting these concepts to our setting,
where the utility function is defined over hybrid domains and models
complex non-linear interactions between attributes, is highly
non-trivial, as our comparisons suggest (see
Section~\ref{sec:soa_comparisons}).  This is an interesting and
challenging direction for future research.


Another research direction is the extension of our approach to handle feedback from multiple DMs~\cite{ALmultipleUsers2011}. In particular, an interesting case study is the exploitation of preferences of previous DMs to minimize the elicitation effort for a new user~\cite{gp2010,PL_multiple_DMs_12}.  
We also plan to extend our algorithm to 
tackle preference drift~\cite{BCEMOprefDrift10}, i.e., the tendency of the DM to change her preferences during the interactive utility elicitation process.
In our combinatorial utility settings, the DM preference drift can be modelled by weights of soft constraints 
evolving over time and by logic formulae gradually changing (e.g., the Boolean term $x_1 \land x_2$ becoming $x_1 \land x_2 \land x_4$ when the DM realizes to have a more complex requirement).

Finally, this paper focused on preference elicitation tasks, involving 
small-scale problems typical of an interaction with a human DM. 
From a more general perspective, CLEO provides a framework for the joint learning and optimization of unknown combinatorial functions, involving both discrete and continuous decision variables.
In principle, when combined with appropriate SMT solvers, CLEO 
could be applied to large combinatorial optimization problems (e.g., arising from industrial applications of combinatorial optimization~\cite{COatWork14}), whose formulation is only \emph{partially} available. 
However, the cost of requiring an explicit
representation of all possible combinations of predicates (even if
limited to the unknown part) would rapidly produce an explosion of
computational and memory requirements. 
An option consists of resorting
to an implicit representation of the function to be optimized, like
the kernelized one we used in~\cite{stochLogicUtFun2010} when learning
quantitative scores. As our previous results seem to
indicate~\cite{stochLogicUtFun2010}, this can produce a degradation in
the quality of returned solutions when the utility function is very
sparse. 
Kernelized versions of zero-norm regularization~\cite{WestEliSch03} could be tried in order to enforce
sparsity in the projected space if needed. Let us however note that the lack
of an explicit formula would prevent the use of all the efficient
refinements of SMT solvers, based on a tight integration between SAT
and theory solvers. A possible alternative is that of pursuing an
incremental feature selection strategy and iteratively solving
increasingly complex approximations of the underlying problem.


\section*{Acknowledgments}

This research was partially supported by grant PRIN 2009LNP494
(Statistical Relational Learning: Algorithms and Applications) from
Italian Ministry of University and Research.


\bibliographystyle{model1-num-names}
\bibliography{paper,chap}


\input{appendix.tex}


\end{document}

%% file: definitions.tex
\newcommand{\lmtvar}[1]{\mbox{\boldmath{$#1$}}}

\newcommand{\lmta}{\lmtvar{A}}
\newcommand{\lmtw}{\lmtvar{w}}
\newcommand{\lmtphi}{\ensuremath{\varphi}}

\newcommand{\lmtPSI}{{\bf \lmtvar{\psi}}}
\newcommand{\lmtxi}{\lmtvar{\xi}}

\def \unsat{\texttt{unsat}}

\def\beq{\begin{equation}}
\def\eeq{\end{equation}}

\def \Bx{{\mathbf x}}

\def \w{{\mathbf w}}
\def \x{{\mathbf x}}

\def \calD{{\cal D}}

\def \be{\begin{equation}}
\def \ee{\end{equation}}

\newtheorem{example}{Example}[section]

\def \Bx{{\mathbf x}}

\DeclareMathOperator*{\argmin}{argmin}
\DeclareMathOperator*{\argmax}{argmax}

\newcommand{\ignore}[1]{}

\newcommand{\ignoreinshort}[1]{\textcolor{green}{#1}} 

\renewcommand{\ignoreinshort}[1]{}

\newcommand{\T}{\ensuremath{\mathcal{T}}\xspace}

\newcommand{\larat}{\ensuremath{\mathcal{LRA}\xspace}}
\newcommand{\laint}{\ensuremath{\mathcal{LIA}\xspace}}

%% file: appendix.tex
\appendix


\section{Additional discussion about the state-of-the-art of preference elicitation}
\label{app:sta_detailed}


This section reviews two notable state-of-the-art approaches for preference elicitation: the body of work adopting the Minimax regret criterion~\cite{regretBased07,Boutilier2010,regretBased06} and the more recent line of research~\cite{softConstr10} developed within the Constraint Programming community. In particular, the latter method shares with CLEO  a constraint-based approach to preference elicitation, resulting in a combinatorial formulation of the DM preferences. 
However, both state-of-the-art methods are not thus suitable for the realistic recommendation tasks considered in our experimental setting, characterized by inaccurate and inconsistent human feedback. In the following, we review these alternative approaches in detail and compare them with CLEO.


\subsection{Minimax regret-based approaches}

The methods developed in the papers~\cite{regretBased07,Boutilier2010,regretBased06} perform preference elicitation under strict uncertainty.  
They assume a parametric formulation of the candidate utility function (hypothesis) in the feasible utility set U. 
The parametrization enables a compact way to specify the feasible set, which is represented by bounds and constrains on the parameters. Uncertainty is thus reduced by tightening the constraints or increasing (decreasing) the lower (upper) bounds.

To make decisions with the partial utility information under strict uncertainty and, in particular, to select the final configuration to be returned to the DM, 
the \emph{minimax regret} decision criterion is used. It prescribes the configuration that minimizes the maximum regret with respect to all the possible realizations of the DM utility function in the set U. Thus, the minimax regret criterion minimizes the \emph{worst-case loss} with respect to the possible realizations of the DM utility function. In detail, the minimax regret criterion is defined in two stages, building on the maximum pairwise regret and the maximum regret.
The maximum pairwise regret of configuration $\Bx$ with respect to configuration $\Bx'$ over the feasible utility set U is defined as:
\begin{equation}
	\label{eq:MPR} 
	\mbox{R}(\Bx,\Bx',\mbox{U})= \max_{u \in \mbox{U}} ~ u(\Bx') - u(\Bx)
\end{equation}
This formulation can be interpreted by assuming an adversary that can
impose any DM utility function $u$ in U and chooses the one 
that maximizes the regret of selecting configuration $\Bx$. The function $u^w = \mbox{arg} \mbox{max } \mbox{R} (\Bx, \Bx', \mbox{U})$ is thus termed the ``adversary's utility'' or ``witness utility''. 
The maximum regret of choosing configuration $\Bx$ with respect to the feasible utility set U is defined as:
\begin{equation}
	\label{eq:MR} 
	\mbox{MR}(\Bx,\mbox{U})= \max_{\Bx'} \mbox{ R}(\Bx,\Bx',\mbox{U})
\end{equation}
Within the ``adversary metaphor'', let us note that the $\Bx'$ chosen by the adversary for the specific $u^w$ is the optimal decision under $u^w$ (i.e., $\Bx'$ maximizes $u^w$) and any alternative choice would give the adversary less utility and thus reduce the user regret.
Finally, the minimax regret of the feasible utility set U is as follows:
\begin{equation}
	\label{eq:MMR} 
	\mbox{MMR} (\mbox{U})= \min_{\Bx} \mbox{ MR}(\Bx,\mbox{U})
\end{equation}
and the configuration $\Bx^r= \mbox{argmin} \mbox{ MR}(\Bx,\mbox{U})$
minimizing the maximum regret is the configuration recommended to the
DM by the minimax regret decision criterion.  The quality of 
configuration $\Bx^r$ is guaranteed to be no more than
$\mbox{MMR}(\mbox{U})$ away from the quality of the DM favourite
configuration, and no alternative configuration has a better guarantee, i.e.,
for all $\Bx \neq \Bx^r$, $\mbox{MR}(\Bx, \mbox{U}) \geq
\mbox{MMR}(\mbox{U})$.


The initial bounds about the utility parameters defined by the DM are
not usually tight enough to identify configurations with provably low
regret, and a configuration satisficing the DM cannot be recommended
without eliciting additional preference information.  This is achieved
through
an interactive elicitation algorithm that asks queries to the DM and,
based on the information elicited, refines the bounds and the
constraints on the utility
parameters. 
The generic framework of the approach is as follows:

\begin{footnotesize}
\begin{pseudo*}
\xn	\xn \textbf{input}: initial constraints (e.g., bounds) on the utility parameters defining  \xl 
\xn	\xn \xn	\xn \xn the initial feasible set U \xl 
 \xl 
	\xn	\xn compute minimax regret $\mbox{MMR}(\mbox{U})$; \xl 
	\xn	\xn \textbf{repeat until} \emph{termination criterion} \xl 
	\xn	\xn \xa  ask query \emph{q}; \xl 
	\xn	\xn \xb  refine U by updating the constraints over utility parameters to reflect the \xl 
	\xn	\xn \xb  response to $\emph{q}$; \xl
	\xn	\xn \xb recompute $\mbox{MMR}(\mbox{U})$ with respect to the refined set U; \xl
	\xn \xn \xc \textbf{return} to the DM the configuration $\Bx^r$ minimizing $\mbox{MR}(\Bx,\mbox{U})$
\end{pseudo*}
\end{footnotesize} 
Computationally tractable techniques have been
proposed~\cite{regretBased07,Boutilier2010,regretBased06} to compute
the minimax regret MMR (U).
The iterative algorithm may be stopped by the DM when she is satisfied by the returned configuration $\Bx^r$ or when the minimax regret $\mbox{MMR}(\mbox{U})$ reaches a certain level $\tau$. When the minimax regret is reduced to the value zero, 
the configuration $\Bx^r$ returned by the algorithm is guaranteed to be the DM favourite configuration.
The minimax regret-based approach also enables a principled method to
define informative queries that will be asked the DM (query selection), and different query strategies have been
proposed~\cite{regretBased07,Boutilier2010,regretBased06}. 



\subsubsection{Comparison with CLEO}
While the CLEO is a preference elicitation method
approximately correct with high probability, 
 the minimax regret-based approaches assume an adversarial entity that     
acts to maximize the DM regret and they aim at beating the adversary 
by recommending the best configuration with respect to the worst case loss. 
However, this adversarial model is not always strongly motivated by real-world
applications, where users are typically interested in the actual
obtained results rather than in regret. 
The main advantage of the regret-based
approaches 
with respect to CLEO is the ability to provide a lower bound
about the quality of the recommended configuration and guarantee the
convergence to provably-optimal results.  However, these theoretical
guarantees are valid under the assumption that the feasible set U
contains the true DM utility function at \emph{any} iteration of the
elicitation process. That is, the regret-based methods do not consider
the uncertain and inconsistent preference information characterizing
the typical human decision processes. As a matter of fact, uncertain
feedback from the DM translates into constraints on the utility
parameters that can potentially rule out the true utility from the
feasible set U. 
Furthermore, the best performance observed in the experiments presented in the paper~\cite{regretBased07} is achieved by query strategies that include   
standard gamble queries, which require the users to state their preference
over a probability distribution of configurations. These queries
 demand a higher DM cognitive load than the comparison queries adopted 
by CLEO, and thus in real-world applications they are more prone
to errors and inconsistent answers from the users.  Without suitable
modifications (e.g., constraints relaxation) to recover from the
inevitable uncertain and inconsistent preference information elicited
from the DM, regret-based approaches cannot be applied in the
realistic problem settings and the noisy test cases that we consider
in this work. %




\subsection{Preference elicitation methods based on constraint satisfaction}
\label{subsec:CPPE}

Recent work in the field of constraint programming~\cite{softConstr10}
shares with CLEO the combinatorial approach to model user
preferences.  It defines the user preferences in terms of
\textit{soft} constraints and introduces constraint optimization
problems where the DM preferences are not completely known before the
solving process starts. Let us first briefly describe the c-semiring
formalism~\cite{semiring97}  
adopted in paper \cite{softConstr10} to model soft constraints.

In soft constraints, a generalization of hard constraints, each
assignment to the variables of one constraint is associated with a
preference value taken from a preference set. The preference value
represents the level of desirability of the assignment to the
variables of the constraint. As the preference score is associated to
a partial assignment to the problem variables, it represents a
\emph{local} preference value.  The desirability of a complete
assignment is defined by a \emph{global} preference score, computed by
applying a combination operator to the local preference values. A set
of soft constraints generates an order (partial or total) over the
complete assignments of the variables of the problem. Given two
solutions of the problem, the preferred one is selected by computing
their global preference
levels. 
Soft constraints are represented by an algebraic structure, called
\textit{c-semiring} (where letter ``c" stays for ``constraint"),
providing two operations for combining ($\times$) and comparing ($+$) 
preference values. In detail, the c-semiring is a tuple $(A, +,
\times, \mathbf 0, \mathbf 1)$ where: 
\begin{itemize}
\item $A$ is a set and $\mathbf 0, \mathbf 1 \in A$;

\item $+$ is commutative, associative and idempotent; $\mathbf 0$ is its unit element and $\mathbf 1$ is its absorbing element;

\item $\times$ is commutative, associative, distributes over $+$; 
$\mathbf 1$ is its unit element and $\mathbf 0$ is its absorbing element.
\end{itemize}
Let us note that a c-semiring is a semiring with additional properties for the two operations: the operation $+$ must be idempotent and with $\mathbf 1$ as absorbing element, the operation $\times$ must be commutative.
The relation $\leq_{A}$ over $A$, $ a_2 \leq_{A} a_1 \mbox{ iff } a_2 + a_1=a_1$, is a partial order, with $\mathbf 0$ and $\mathbf 1$ its minimum and maximum elements, respectively. 
The relation $\leq_{A}$ allows to compare (some of) the desirability levels, with $ a_2 \leq_{A} a_1$ meaning that $a_1$ is ``better" than $a_2$; $\mathbf 0$ and $\mathbf 1$ represent the worst and the best preference levels, respectively, and the operations $+$ and $\times$ are monotone on $\leq_{A}$. 
Consider, e.g., the following instance of c-semiring:
$$(\{5,10,15, \dots , 50 \}, \mbox{max}, \mbox{min}, 5, 50)$$
with preference values from the set $\{5,10,15, \dots , 50 \}$ and elements $\mathbf 0$ and $\mathbf 1$ represented by the values $5$ and $50$, respectively.
The desirability of a complete assignment is obtained by taking its minimum local preference value. A complete assignment $c_1$ with preference score $a_1$ is preferred to a complete assignment $c_2$ with lower preference score $a_2$. That is, $ a_2 \leq_{A} a_1 \mbox{ iff max} (a_2, a_1)= a_1$.

The generality of the semiring-based soft constraint formalism permits
to express several kinds of preferences, including partially ordered
ones. For example, different instances of c-semirings encode weighted
or probabilistic soft constraint satisfaction
problems~\cite{bipolarPrefs10}. However, the c-semiring formalism can
model just \textit{negative} preferences. First, the best element in
the ordering induced by $\leq_{A}$, denoted by $\mathbf 1$, behaves as
indifference, since $\forall a \in A, 1 \times a=a$. This result is
consistent with intuition: when using only negative preferences,
indifference is the best level of desirability that can be
expressed. Furthermore, the combination of desirability levels returns
a lower overall preference, since $a \times b \leq_{A} a,b$, again
consistently with the fact of dealing with negative preferences.

Preference elicitation strategies have been
introduced~\cite{softConstr10} within this formalism in order to deal
with scenarios where preference value information is partially
unknown. Some of the local preference values attached to soft
constraints are assumed to be missing, and the DM is asked for an
explicit feedback on specific assignments for these constraints, in
terms of score values quantifying her preference for a certain
assignment. The elicitation strategy is aimed at minimizing the number
of queries to the DM.

\subsubsection{Comparison with CLEO}

Concerning expressivity of the representation formalisms, the work
in~\cite{semiringMaxSat07} shows how to encode semiring-based soft
constraint satisfaction problem (SCSP) instances into equivalent
weighted MAX-SAT formulations. Each solution of the latter instance
corresponds to a solution of the former one. Details on the encoding
algorithm can be found in~\ref{app:SCSP_into_MAX_SAT}.  The rationale for the MAX-SAT
encoding is the exploitation of the efficient and widely studied
techniques implemented in modern SAT solvers, which can efficiently
handle large-size structured problems~\cite{sat_solvers_08}.  The
encoding can in principle be applied also to SCSPs with continuous
decision variables or discrete variables defined over large size
finite domains, possibly however at the cost of a significant blow-up
in the translation. In this case, one may cast the SCSP instance into
a weighted MAX-SMT rather than a weighted MAX-SAT formulation.  

Concerning the preference elicitation setting, our formulation assumes
a much more limited amount of initial knowledge about the problem to
be optimized. In the work on preference elicitation for
SCSPs~\cite{softConstr10}, decision variables, soft constraint
topology and structure are assumed to be known in advance and the
incomplete initial information consists only of missing local
preference values. 
CLEO assumes complete ignorance about the structure of the constraints
over the decisional variables of the user.
The initial problem knowledge is limited to a set of catalog
attributes. CLEO extracts the decisional items of the DM from
the set of catalog attributes and learns the weighted constraints constructed
from them modeling the DM preferences.  If the MAX-SAT encoding is
applied to the SCSP with missing preferences, it produces a Boolean
formula where some of the weights of the terms are not known. On the
other hand, CLEO handles MAX-SAT
instances where both the constraints and their associated weights are
initially unknown and are learnt by interacting with the DM.


Furthermore, the technique in~\cite{softConstr10} is based on \textit{local} elicitation queries, with the final user asked to reveal her preferences about assignments for specific soft constraints. \textit{Global} preferences or bounds for global preferences associated to complete solutions of the problem are derived from the local preference information. 
CLEO goes in the opposite direction: it asks the user to compare complete solutions and learns local utilities (i.e., the weights of the constraints of the logic formula) from global preference values. 
In many cases, recognizing appealing or unsatisfactory global solutions may be much easier than defining local utility functions, associated to partial solutions. For example, while scheduling a set of activities, the evaluation of complete schedules may be more affordable than assessing how specific ordering choices between couples of activities contribute to the global preference value.
Furthermore the preference elicitation technique in~\cite{softConstr10} asks the DM for quantitative evaluations of partial solutions: she does not just rank couples of activities, she provides score values quantifying her preference for the partial activity rankings, a much more demanding task. 

In order to reduce the embarrassment of the decision maker when
specifying precise preference scores, interval-valued
constraints~\cite{intervalConstr10} allow users to state an interval
of utility values for each instantiation of the variables of a
constraint.  
As a matter of fact, the informal definitions of
degrees of preference such as ``quite high", ``more or less", ``low"
or ``undesirable" cannot be naturally mapped to precise preference
scores.  However, the technique described in~\cite{intervalConstr10}
requires the user to provide all the information she has about the
problem (in terms of preference intervals) \textit{before} the solving
phase, without seeing any optimization result.

Even if interval-valued constraints~\cite{intervalConstr10} have been
introduced to handle uncertainty in the evaluations of the DM,
inconsistent preference information is not
addressed~\cite{softConstr10}. This is a requirement to retain the
optimality guarantees provided by the preference elicitation
strategy.  
Conversely, CLEO 
trades optimality for robustness and can effectively deal with
imprecise information from the DM, modelled in terms of inaccurate
ranking of the candidate solutions.


Finally, while the work in~\cite{softConstr10} considers
\textit{unipolar} preference problems, modeling just negative
preferences, CLEO naturally accounts for bipolar preference
problems, with the final user specifying what she likes and what she
dislikes. Bipolar preference problems provide a better representation
of the typical human decision process, where the degree of preference
for a solution reflects the compensation value obtained by comparing
its advantages with the disadvantages. Let us note that the work
in~\cite{bipolarPrefs10} extends the soft constraint formalism to
account for bipolar preference problems.


\subsubsection{Econding SCSP into weighted MAX-SAT instances}
\label{app:SCSP_into_MAX_SAT}

The work in~\cite{semiringMaxSat07} introduces a method to encode a semiring-based soft constraint satisfaction problem (SCSP) instance into a weighted MAX-SAT instance, with
each solution of the generated MAX-SAT instance corresponding to a solution of the original SCSP.
With no loss of generality, assume a soft constraint problem with $n$ variables $v_1, \dots v_n$ having domain $D_1, \dots D_n$, and $m$ constraints $c_1, \dots c_m$. Each instantiation of the variables of a constraint $c_j \mbox{, } j=1 \dots m \mbox{, }$ is associated with a value from the c-semiring $(A, +, \times, \mathbf 0, \mathbf 1)$.
For each variable $v_i$, $i=1 \dots n$, and each value $d \in D_i$, a Boolean variable $b_{i,d}$ is introduced. When $b_{i,d}$ is set to true then $v_i$ is assigned the value $d \in D_i$. 
The variables $b_{i,d}$, $i=1 \dots n$, $d \in D_i$, represent the Boolean variables of the weighted MAX-SAT problem.

The set of Boolean constraints of the MAX-SAT problem consists of clauses ensuring that each variable $v_i$, $i= 1 \dots n$, is assigned exactly one value $d \in D_i$, 
 and of terms representing the soft constraints of the original SCSP.
In the former case, for each variable $v_i, i=1 \dots n$, the \emph{at-least-one-value} hard clause:
$$ (b_{i, d_1} \lor b_{i, d_2} \lor \dots \lor b_{i, d_{|D_i|}}) $$
and the set of $(|D_i| \dot (D_i| -1)) /2$ binary \emph{at-max-one-value} hard clauses:
$$ (\lnot b_{i, d_j} \lor \lnot b_{i, d_k}) \mbox{ for every pair } (d_j, d_k) \mbox{ with } d_j,d_k \in D_i \mbox{ and } 1\leq j < k \leq |D_i|$$ 
are generated. They ensure that for each $i \in \{1 \dots n\}$ exactly one variable $b_{i,j}$, $j \in \{1,2, \dots, |D_i|\}$ is set to true.

Each soft constraint of the original SCSP is represented by a set of weighted Boolean terms encoding all the possible assignments of values (i.e., configurations) to its variables. The weight of a term is set to the c-semiring value associated to the encoded  configuration. 
For example, consider a binary soft constraint over variables $v_1$ and $v_2$ both with discrete domain $D=\{ 1,2,3 \}$ and with preference scores defined by the semiring 
$(\{5,10,15, \dots , 50 \},$ $\mbox{max}, \mbox{min}, 5, 50)$.
The possible configurations are specified in Table~\ref{tab:softConMappingExa} (left).
Each row shows an assignment of values to $v_1$ and $v_2$ and the c-semiring value associated to the assignment. Given the six Boolean variables $b_{1,d}$ and $b_{2,d}$ with $d=1,2,3$ defined as above, the soft constraint in Table~\ref{tab:softConMappingExa} (left) is encoded into the set of Boolean terms in Table~\ref{tab:softConMappingExa} (right). 
 
\begin{table}
	\centering
	\par\vspace{1ex}

	\begin{minipage}[t]{0.4\textwidth}

		\begin{tabular}{c c c} 
		\hline\hline 
		$v_1$  & $v_2$ & preference value \\ [0.5ex] 
		\hline 
		1    &  1  &   10\\        
		1    &  2  &   40\\         
		1    &  3  &   50\\
		2    &  1  &   5\\        
		2    &  2  &   10\\         
		2    &  3  &   30\\
		3    &  1  &   5\\        
		3    &  2  &   5\\
		3    &  3  &   10 \\ [1ex] 
		\hline 
		\end{tabular}

	\end{minipage}
	\begin{minipage}[t]{0.4\textwidth}
		\begin{tabular}{c c c} 
		\hline\hline 
		num & term & weight \\ [0.5ex] 
		\hline 
		1 & $(b_{1,1} \land  b_{2,1})$  &   10\\        
		2 & $(b_{1,1} \land  b_{2,2})$  &   40\\         
		3 & $(b_{1,1} \land  b_{2,3})$  &   50\\
		4 & $(b_{1,2} \land  b_{2,1})$  &   5\\        
		5 & $(b_{1,2} \land  b_{2,2})$  &   10\\         
		6 & $(b_{1,2} \land  b_{2,3})$  &   30\\
		7 & $(b_{1,3} \land  b_{2,1})$  &   5\\        
		8 & $(b_{1,3} \land  b_{2,2})$  &   5\\
		9 & $(b_{1,3} \land  b_{2,3})$  &   10 \\ [1ex] 
		\hline 
		\end{tabular}
	\end{minipage}
\caption{\label{tab:softConMappingExa}(left) example of soft constraint. The DM prefers assignments with $v_1 < v_2$. (Right) weighted Boolean terms encoding the soft constraint defined in the left table. When the Boolean variable $b_{i,d}  : d \in D_i$, is set to true then $v_i$ is assigned the value $d$.}
\end{table}

A structured MAX-SAT formulation can be obtained by considering generalized Boolean clauses which are the disjunction of the terms encoding for a given soft constraint the assignments with the same preference value.
For example, the terms defined at rows number $1,5,9$ in Table~\ref{tab:softConMappingExa} (right) can be merged into a single generalized weighted clause:
$$
(b_{1,1} \land  b_{2,1})  \lor (b_{1,2} \land  b_{2,2}) \lor  (b_{1,3} \land  b_{2,3})
$$
with weight equal to $10$.
Furthermore, each \emph{at-least-one-value} and \emph{at-max-one-value} hard clause $h$ can be cast into a soft clause represented by its negation $\lnot h$ and with associated the semiring value $\mathbf 0$~\cite{semiringMaxSat07}. The value $\mathbf 0$ is indeed both the minimum value in the partial order defined by the relation $\leq_{A}$ and the absorbing element for the operator $\times$ combining the semiring values. Therefore, a candidate solution $\mathbf b$ of the generated MAX-SAT instance that does not satisfy one of these soft clauses receives the minimum semiring value $\mathbf 0$.
However, this implementation of the hard clauses does not allow to discern infeasible solutions from feasible ones with lowest possible preference, i.e., feasible solutions  getting the lowest semiring value. 

Given the generated MAX-SAT formulation, the optimization task consists of finding the assignment $\mathbf b^*$ to the Boolean variables $b_{i,d}$, $i=1 \dots n$, $d \in D_i$, maximizing $f(\mathbf b)$, with $f(\mathbf b)$ the semiring value obtained by combining by the operator $\times$ the weights of the solution components satisfied by $\mathbf b$. 
Each candidate solution ($\mathbf b$, $f(\mathbf b)$) of the generated MAX-SAT instance identifies an assignment of values to the variables $v_i \mbox{, } i=1 \dots n,$ of the original SCSP with associated semiring value $f(\mathbf b)$.

%% file: CLEO_paper.bbl
\begin{thebibliography}{45}
\expandafter\ifx\csname natexlab\endcsname\relax\def\natexlab#1{#1}\fi
\providecommand{\bibinfo}[2]{#2}
\ifx\xfnm\relax \def\xfnm[#1]{\unskip,\space#1}\fi
\bibitem[{Peintner et~al.(2008)Peintner, Viappiani, and
  Yorke-Smith}]{pe_survey08}
\bibinfo{author}{B.~Peintner}, \bibinfo{author}{P.~Viappiani},
  \bibinfo{author}{N.~Yorke-Smith},
\newblock \bibinfo{title}{{Preferences in Interactive Systems: Technical
  Challenges and Case Studies}},
\newblock \bibinfo{journal}{AI Magazine} \bibinfo{volume}{29}
  (\bibinfo{year}{2008}) \bibinfo{pages}{13--24}.
\bibitem[{March(1978)}]{Bounded_ration_78}
\bibinfo{author}{J.~G. March},
\newblock \bibinfo{title}{{Bounded Rationality, Ambiguity, and the Engineering
  of Choice}},
\newblock \bibinfo{journal}{The Bell Journal of Economics} \bibinfo{volume}{9}
  (\bibinfo{year}{1978}) \bibinfo{pages}{pp. 587--608}.
\bibitem[{Guo and Sanner(2010)}]{GSMalgorithm10}
\bibinfo{author}{S.~Guo}, \bibinfo{author}{S.~Sanner},
\newblock \bibinfo{title}{{Real-time Multiattribute Bayesian Preference
  Elicitation with Pairwise Comparison Queries}},
\newblock \bibinfo{journal}{Journal of Machine Learning Research - Proceedings
  Track} \bibinfo{volume}{9} (\bibinfo{year}{2010}) \bibinfo{pages}{289--296}.
\bibitem[{Braziunas and Boutilier(2007)}]{regretBased07}
\bibinfo{author}{D.~Braziunas}, \bibinfo{author}{C.~Boutilier},
\newblock \bibinfo{title}{Minimax regret based elicitation of generalized
  additive utilities},
\newblock in: \bibinfo{booktitle}{Proceedings of the Twenty-third Conference on
  Uncertainty in Artificial Intelligence (UAI-07)},
  \bibinfo{address}{Vancouver}, pp. \bibinfo{pages}{25--32}.
\bibitem[{Boutilier et~al.(2010)Boutilier, Regan, and
  Viappiani}]{Boutilier2010}
\bibinfo{author}{C.~Boutilier}, \bibinfo{author}{K.~Regan},
  \bibinfo{author}{P.~Viappiani},
\newblock \bibinfo{title}{{Simultaneous Elicitation of Preference Features and
  Utility}},
\newblock in: \bibinfo{booktitle}{Proceedings of the Twenty-fourth AAAI
  Conference on Artificial Intelligence (AAAI-10)}, \bibinfo{publisher}{AAAI
  press}, \bibinfo{address}{Atlanta, GA, USA}, \bibinfo{year}{2010}, pp.
  \bibinfo{pages}{1160--1167}.
\bibitem[{Boutilier et~al.(2006)Boutilier, Patrascu, Poupart, and
  Schuurmans}]{regretBased06}
\bibinfo{author}{C.~Boutilier}, \bibinfo{author}{R.~Patrascu},
  \bibinfo{author}{P.~Poupart}, \bibinfo{author}{D.~Schuurmans},
\newblock \bibinfo{title}{{Constraint-based Optimization and Utility
  Elicitation using the Minimax Decision Criterion}},
\newblock \bibinfo{journal}{Artificial Intelligence} \bibinfo{volume}{170}
  (\bibinfo{year}{2006}) \bibinfo{pages}{686--713}.
\bibitem[{Bonilla et~al.(2010)Bonilla, Guo, and Sanner}]{gp2010}
\bibinfo{author}{E.~Bonilla}, \bibinfo{author}{S.~Guo},
  \bibinfo{author}{S.~Sanner},
\newblock \bibinfo{title}{{Gaussian Process Preference Elicitation}},
\newblock in: \bibinfo{editor}{J.~Lafferty}, \bibinfo{editor}{C.~K.~I.
  Williams}, \bibinfo{editor}{J.~Shawe-Taylor}, \bibinfo{editor}{R.~Zemel},
  \bibinfo{editor}{A.~Culotta} (Eds.), \bibinfo{booktitle}{Advances in Neural
  Information Processing Systems 23: 24th Annual Conference on Neural
  Information Processing Systems}, \bibinfo{year}{2010}, pp.
  \bibinfo{pages}{262--270}.
\bibitem[{Viappiani(2012)}]{MC12}
\bibinfo{author}{P.~Viappiani},
\newblock \bibinfo{title}{{Monte Carlo Methods for Preference Learning}},
\newblock in: \bibinfo{booktitle}{Proceedings of the 6th Learning and
  Intelligent OptimizatioN Conference (LION VI)}, LNCS,
  \bibinfo{publisher}{Springer Verlag}, \bibinfo{address}{Paris, France},
  \bibinfo{year}{2012}.
\bibitem[{Birlutiu et~al.(2012)Birlutiu, Groot, and
  Heskes}]{PL_multiple_DMs_12}
\bibinfo{author}{A.~Birlutiu}, \bibinfo{author}{P.~Groot},
  \bibinfo{author}{T.~Heskes},
\newblock \bibinfo{title}{Efficiently learning the preferences of people},
\newblock \bibinfo{journal}{Machine Learning}  (\bibinfo{year}{2012})
  \bibinfo{pages}{1--28}.
\bibitem[{Gelain et~al.(2010)Gelain, Pini, Rossi, Venable, and
  Walsh}]{softConstr10}
\bibinfo{author}{M.~Gelain}, \bibinfo{author}{M.~S. Pini},
  \bibinfo{author}{F.~Rossi}, \bibinfo{author}{K.~B. Venable},
  \bibinfo{author}{T.~Walsh},
\newblock \bibinfo{title}{{Elicitation Strategies for Soft Constraint Problems
  with Missing Preferences: Properties, Algorithms and Experimental Studies}},
\newblock \bibinfo{journal}{Artificial Intelligence Journal}
  \bibinfo{volume}{174} (\bibinfo{year}{2010}) \bibinfo{pages}{270--294}.
\bibitem[{Nieuwenhuis and Oliveras(2006)}]{NieOli06}
\bibinfo{author}{R.~Nieuwenhuis}, \bibinfo{author}{A.~Oliveras},
\newblock \bibinfo{title}{{On SAT Modulo Theories and Optimization Problems}},
\newblock in: \bibinfo{booktitle}{Theory and Applications of Satisfiability
  Testing}, LNCS, \bibinfo{publisher}{Springer}, \bibinfo{year}{2006}, pp.
  \bibinfo{pages}{156--169}.
\bibitem[{Teso et~al.(2015)Teso, Sebastiani, and Passerini}]{Teso2015}
\bibinfo{author}{S.~Teso}, \bibinfo{author}{R.~Sebastiani},
  \bibinfo{author}{A.~Passerini},
\newblock \bibinfo{title}{Structured learning modulo theories},
\newblock \bibinfo{journal}{Artificial Intelligence}  (\bibinfo{year}{2015}).
\bibitem[{Miller(1956)}]{mil56}
\bibinfo{author}{G.~A. Miller},
\newblock \bibinfo{title}{{The Magical Number Seven, Plus or Minus Two: Some
  Limits on Our Capacity for Processing Information}},
\newblock \bibinfo{journal}{The Psychological Review} \bibinfo{volume}{63}
  (\bibinfo{year}{1956}) \bibinfo{pages}{81--97}.
\bibitem[{Tibshirani(1996)}]{Tib96}
\bibinfo{author}{R.~Tibshirani},
\newblock \bibinfo{title}{{Regression Shrinkage and Selection Via the Lasso}},
\newblock \bibinfo{journal}{Journal of the Royal Statistical Society, Series B}
  \bibinfo{volume}{58} (\bibinfo{year}{1996}) \bibinfo{pages}{267--288}.
\bibitem[{Campigotto et~al.(2011)Campigotto, Passerini, and
  Battiti}]{stochLogicUtFun2010}
\bibinfo{author}{P.~Campigotto}, \bibinfo{author}{A.~Passerini},
  \bibinfo{author}{R.~Battiti},
\newblock \bibinfo{title}{{Active Learning of Combinatorial Features for
  Interactive Optimization}},
\newblock in: \bibinfo{booktitle}{Proceedings of the 5th Learning and
  Intelligent OptimizatioN Conference (LION V), Rome, Italy, Jan 17-21, 2011},
  LNCS, \bibinfo{publisher}{Springer Verlag}, \bibinfo{year}{2011}.
\bibitem[{Barrett et~al.(2009)Barrett, Sebastiani, Seshia, and
  Tinelli}]{BarSebSes09}
\bibinfo{author}{C.~Barrett}, \bibinfo{author}{R.~Sebastiani},
  \bibinfo{author}{S.~A. Seshia}, \bibinfo{author}{C.~Tinelli},
\newblock \bibinfo{title}{{Satisfiability Modulo Theories}},
\newblock in: \bibinfo{booktitle}{Handbook of Satisfiability},
  \bibinfo{publisher}{IOS Press}, \bibinfo{year}{2009}, pp.
  \bibinfo{pages}{825--885}.
\bibitem[{Sebastiani(2007)}]{sebastiani07}
\bibinfo{author}{R.~Sebastiani},
\newblock \bibinfo{title}{{Lazy Satisfiability Modulo Theories}},
\newblock \bibinfo{journal}{Journal on Satisfiability, Boolean Modeling and
  Computation, JSAT} \bibinfo{volume}{3} (\bibinfo{year}{2007})
  \bibinfo{pages}{141--224}.
\bibitem[{Barrett et~al.(2009)Barrett, Sebastiani, Seshia, and
  Tinelli}]{BSST09HBSAT}
\bibinfo{author}{C.~Barrett}, \bibinfo{author}{R.~Sebastiani},
  \bibinfo{author}{S.~A. Seshia}, \bibinfo{author}{C.~Tinelli},
  \bibinfo{title}{Satisfiability Modulo Theories}, Frontiers in Artificial
  Intelligence and Applications, \bibinfo{publisher}{IOS Press}, pp.
  \bibinfo{pages}{825--885}.
\bibitem[{Nieuwenhuis and Oliveras(2006)}]{nieuwenhuis_sat06}
\bibinfo{author}{R.~Nieuwenhuis}, \bibinfo{author}{A.~Oliveras},
\newblock \bibinfo{title}{{On SAT Modulo Theories and Optimization Problems}},
\newblock in: \bibinfo{booktitle}{Proc. Theory and Applications of
  Satisfiability Testing - SAT 2006}, volume \bibinfo{volume}{4121} of
  \textit{\bibinfo{series}{Lecture Notes in Computer Science}},
  \bibinfo{publisher}{Springer}, \bibinfo{year}{2006}.
\bibitem[{Cimatti et~al.(2010)Cimatti, Franz{\'e}n, Griggio, Sebastiani, and
  Stenico}]{cimattifgss10}
\bibinfo{author}{A.~Cimatti}, \bibinfo{author}{A.~Franz{\'e}n},
  \bibinfo{author}{A.~Griggio}, \bibinfo{author}{R.~Sebastiani},
  \bibinfo{author}{C.~Stenico},
\newblock \bibinfo{title}{Satisfiability modulo the theory of costs:
  Foundations and applications},
\newblock in: \bibinfo{booktitle}{Proc. Tools and Algorithms for the
  Construction and Analysis of Systems, TACAS}, volume \bibinfo{volume}{6015}
  of \textit{\bibinfo{series}{Lecture Notes in Computer Science}},
  \bibinfo{publisher}{Springer}, \bibinfo{year}{2010}, pp.
  \bibinfo{pages}{99--113}.
\bibitem[{Cimatti et~al.(2013)Cimatti, Griggio, Schaafsma, and
  Sebastiani}]{cgss_sat13_maxsmt}
\bibinfo{author}{A.~Cimatti}, \bibinfo{author}{A.~Griggio},
  \bibinfo{author}{B.~J. Schaafsma}, \bibinfo{author}{R.~Sebastiani},
\newblock \bibinfo{title}{{A Modular Approach to MaxSAT Modulo Theories}},
\newblock in: \bibinfo{booktitle}{International Conference on Theory and
  Applications of Satisfiability Testing, SAT}, volume \bibinfo{volume}{7962}
  of \textit{\bibinfo{series}{Lecture Notes in Computer Science}},
  \bibinfo{publisher}{Springer}, \bibinfo{year}{2013}.
\bibitem[{Joachims(2002)}]{ranksvm}
\bibinfo{author}{T.~Joachims},
\newblock \bibinfo{title}{Optimizing search engines using clickthrough data},
\newblock in: \bibinfo{booktitle}{Proceedings of the Eighth ACM SIGKDD
  International Conference on Knowledge Discovery and Data Mining}, KDD '02,
  \bibinfo{publisher}{ACM}, \bibinfo{address}{New York, NY, USA},
  \bibinfo{year}{2002}, pp. \bibinfo{pages}{133--142}.
\bibitem[{Friedman et~al.(2004)Friedman, Hastie, Rosset, and
  Tibshirani}]{Friedman04}
\bibinfo{author}{J.~Friedman}, \bibinfo{author}{T.~Hastie},
  \bibinfo{author}{S.~Rosset}, \bibinfo{author}{R.~Tibshirani},
\newblock \bibinfo{title}{Discussion of boosting papers},
\newblock \bibinfo{journal}{Annals of Statistics} \bibinfo{volume}{32}
  (\bibinfo{year}{2004}) \bibinfo{pages}{102--107}.
\bibitem[{Pu and Chen(2008)}]{PuChen2008}
\bibinfo{author}{P.~Pu}, \bibinfo{author}{L.~Chen},
\newblock \bibinfo{title}{{User-Involved Preference Elicitation for Product
  Search and Recommender Systems}},
\newblock \bibinfo{journal}{AI magazine} \bibinfo{volume}{29}
  (\bibinfo{year}{2008}) \bibinfo{pages}{93--103}.
\bibitem[{Braziunas(2006)}]{Braziunas06techre}
\bibinfo{author}{D.~Braziunas}, \bibinfo{title}{{Computational Approaches to
  Preference Elicitation}}, \bibinfo{type}{Technical Report}, Department of
  Computer Science, University of Toronto, \bibinfo{year}{2006}.
\bibitem[{Domshlak et~al.(2011)Domshlak, H\"{u}llermeier, Kaci, and
  Prade}]{PEoverview11}
\bibinfo{author}{C.~Domshlak}, \bibinfo{author}{E.~H\"{u}llermeier},
  \bibinfo{author}{S.~Kaci}, \bibinfo{author}{H.~Prade},
\newblock \bibinfo{title}{{Preferences in {AI}: An overview}},
\newblock \bibinfo{journal}{Artificial Intelligence} \bibinfo{volume}{175}
  (\bibinfo{year}{2011}) \bibinfo{pages}{1037--1052}.
\bibitem[{Savage(1951)}]{Savage1951}
\bibinfo{author}{L.~J. Savage},
\newblock \bibinfo{title}{{The Theory of Statistical Decision}},
\newblock \bibinfo{journal}{Journal of the American Statistical Association}
  \bibinfo{volume}{46} (\bibinfo{year}{1951}) \bibinfo{pages}{55--67}.
\bibitem[{Weng and Lin(2011)}]{rank_algo_11}
\bibinfo{author}{R.~C. Weng}, \bibinfo{author}{C.-J. Lin},
\newblock \bibinfo{title}{{A Bayesian Approximation Method for Online
  Ranking}},
\newblock \bibinfo{journal}{Journal of Machine Learning Research}
  \bibinfo{volume}{12} (\bibinfo{year}{2011}) \bibinfo{pages}{267--300}.
\bibitem[{Tsukida and Gupta(2011)}]{pairwiseComp11}
\bibinfo{author}{K.~Tsukida}, \bibinfo{author}{M.~R. Gupta},
  \bibinfo{title}{{How to Analyze Paired Comparison Data}},
  \bibinfo{type}{Technical Report} \bibinfo{number}{No. UWEETR-2011-004},
  Washington University, Dep. of Electrical Engineering, Seattle, USA,
  \bibinfo{year}{May 2011}. \bibinfo{note}{(as of June 2015)}.
\bibitem[{Mcfadden(2001)}]{economicchoices01}
\bibinfo{author}{D.~Mcfadden},
\newblock \bibinfo{title}{{Economic Choices}},
\newblock \bibinfo{journal}{American Economic Review} \bibinfo{volume}{91}
  (\bibinfo{year}{2001}) \bibinfo{pages}{351--378}.
\bibitem[{Dutertre and de~Moura(2006)}]{yices_paper}
\bibinfo{author}{B.~Dutertre}, \bibinfo{author}{L.~de~Moura},
\newblock \bibinfo{title}{{A Fast Linear-Arithmetic Solver for DPLL(T)}},
\newblock in: \bibinfo{booktitle}{{Proceedings of the 18th Computer-Aided
  Verification conference}}, LNCS, \bibinfo{publisher}{Springer},
  \bibinfo{year}{2006}, pp. \bibinfo{pages}{81--94}.
\bibitem[{Chakrabarti et~al.(2008)Chakrabarti, Khanna, Sawant, and
  Bhattacharyya}]{ChaetAl08}
\bibinfo{author}{S.~Chakrabarti}, \bibinfo{author}{R.~Khanna},
  \bibinfo{author}{U.~Sawant}, \bibinfo{author}{C.~Bhattacharyya},
\newblock \bibinfo{title}{Structured learning for non-smooth ranking losses},
\newblock in: \bibinfo{booktitle}{14th ACM SIGKDD international conference on
  Knowledge discovery and data mining}, KDD '08, \bibinfo{publisher}{ACM},
  \bibinfo{year}{2008}, pp. \bibinfo{pages}{88--96}.
\bibitem[{Settles(2009)}]{Set09}
\bibinfo{author}{B.~Settles}, \bibinfo{title}{{Active Learning Literature
  Survey}}, \bibinfo{type}{Technical Report} \bibinfo{number}{Computer Sciences
  Technical Report 1648}, University of Wisconsin-Madison,
  \bibinfo{year}{2009}.
\bibitem[{Branke et~al.(2008)Branke, Deb, Miettinen, and
  S{\l}owi{\'{n}}ski}]{MooInt08}
\bibinfo{editor}{J.~Branke}, \bibinfo{editor}{K.~Deb},
  \bibinfo{editor}{K.~Miettinen}, \bibinfo{editor}{R.~S{\l}owi{\'{n}}ski}
  (Eds.), \bibinfo{title}{{Multiobjective Optimization: Interactive and
  Evolutionary Approaches}}, \bibinfo{publisher}{Springer Verlag},
  \bibinfo{year}{2008}.
\bibitem[{Radlinski and Joachims(2007)}]{RadJoa07}
\bibinfo{author}{F.~Radlinski}, \bibinfo{author}{T.~Joachims},
\newblock \bibinfo{title}{{Active exploration for learning rankings from
  clickthrough data}},
\newblock in: \bibinfo{booktitle}{13th ACM SIGKDD international conference on
  Knowledge discovery and data mining (KDD '07)}, \bibinfo{publisher}{{ACM
  Press}}, \bibinfo{year}{2007}, pp. \bibinfo{pages}{570--579}.
\bibitem[{Xu et~al.(2007)Xu, Akella, and Zhang}]{DiversityAndDensity07}
\bibinfo{author}{Z.~Xu}, \bibinfo{author}{R.~Akella},
  \bibinfo{author}{Y.~Zhang},
\newblock \bibinfo{title}{{Incorporating Diversity and Density in Active
  Learning for Relevance Feedback}},
\newblock in: \bibinfo{editor}{G.~Amati}, \bibinfo{editor}{C.~Carpineto},
  \bibinfo{editor}{G.~Romano} (Eds.), \bibinfo{booktitle}{Advances in
  Information Retrieval}, volume \bibinfo{volume}{4425} of
  \textit{\bibinfo{series}{LNCS}}, \bibinfo{publisher}{Springer},
  \bibinfo{year}{2007}, pp. \bibinfo{pages}{246--257}.
\bibitem[{Yan et~al.(2011)Yan, Rosales, Fung, and Dy}]{ALmultipleUsers2011}
\bibinfo{author}{Y.~Yan}, \bibinfo{author}{R.~Rosales},
  \bibinfo{author}{G.~Fung}, \bibinfo{author}{J.~Dy},
\newblock \bibinfo{title}{{Active Learning from Crowds}},
\newblock in: \bibinfo{editor}{L.~Getoor}, \bibinfo{editor}{T.~Scheffer}
  (Eds.), \bibinfo{booktitle}{Proceedings of the 28th International Conference
  on Machine Learning (ICML-11)}, \bibinfo{publisher}{ACM},
  \bibinfo{address}{New York, NY, USA}, \bibinfo{year}{2011}, pp.
  \bibinfo{pages}{1161--1168}.
\bibitem[{Campigotto et~al.(2010)Campigotto, Passerini, and
  Battiti}]{BCEMOprefDrift10}
\bibinfo{author}{P.~Campigotto}, \bibinfo{author}{A.~Passerini},
  \bibinfo{author}{R.~Battiti},
\newblock \bibinfo{title}{Handling concept drift in preference learning for
  interactive decision making},
\newblock in: \bibinfo{booktitle}{Online proceedings of the 1st International
  Workshop on Handling Concept Drift in Adaptive Information Systems (HaCDAIS
  2010), Barcelona, Spain, Sept 24, 2010.}
\bibitem[{Paschos(2014)}]{COatWork14}
\bibinfo{author}{V.~T. Paschos}, \bibinfo{title}{{Applications of combinatorial
  optimization}}, \bibinfo{publisher}{Mathematics and Statistics Series,
  Wiley-ISTE, 2nd ed.}, \bibinfo{year}{2014}.
\bibitem[{Weston et~al.(2003)Weston, Elisseeff, Sch\"{o}lkopf, and
  Tipping}]{WestEliSch03}
\bibinfo{author}{J.~Weston}, \bibinfo{author}{A.~Elisseeff},
  \bibinfo{author}{B.~Sch\"{o}lkopf}, \bibinfo{author}{M.~Tipping},
\newblock \bibinfo{title}{Use of the zero norm with linear models and kernel
  methods},
\newblock \bibinfo{journal}{{Journal of Machine Learning Research}}
  \bibinfo{volume}{3} (\bibinfo{year}{2003}) \bibinfo{pages}{1439--1461}.
\bibitem[{Bistarelli et~al.(1997)Bistarelli, Montanari, and Rossi}]{semiring97}
\bibinfo{author}{S.~Bistarelli}, \bibinfo{author}{U.~Montanari},
  \bibinfo{author}{F.~Rossi},
\newblock \bibinfo{title}{{Semiring-based Constraint Solving and
  Optimization}},
\newblock \bibinfo{journal}{Journal of {ACM}} \bibinfo{volume}{44}
  (\bibinfo{year}{1997}) \bibinfo{pages}{201--236}.
\bibitem[{Bistarelli et~al.(2010)Bistarelli, Pini, Rossi, and
  Venable}]{bipolarPrefs10}
\bibinfo{author}{S.~Bistarelli}, \bibinfo{author}{M.~S. Pini},
  \bibinfo{author}{F.~Rossi}, \bibinfo{author}{K.~B. Venable},
\newblock \bibinfo{title}{From soft constraints to bipolar preferences:
  modelling framework and solving issues},
\newblock \bibinfo{journal}{Journal of Experimental and Theoretical Artificial
  Intelligence} \bibinfo{volume}{22} (\bibinfo{year}{2010})
  \bibinfo{pages}{135--158}.
\bibitem[{Leenen et~al.(2007)Leenen, Anbulagan, Meyer, and
  Ghose}]{semiringMaxSat07}
\bibinfo{author}{L.~Leenen}, \bibinfo{author}{Anbulagan},
  \bibinfo{author}{T.~Meyer}, \bibinfo{author}{A.~K. Ghose},
\newblock \bibinfo{title}{{Modeling and Solving Semiring Constraint
  Satisfaction Problems by Transformation to Weighted Semiring Max-SAT}},
\newblock in: \bibinfo{booktitle}{20th Australian Joint Conference on
  Artificial Intelligence}, volume \bibinfo{volume}{4830} of
  \textit{\bibinfo{series}{LNCS}}, \bibinfo{publisher}{Springer},
  \bibinfo{year}{2007}, pp. \bibinfo{pages}{202--212}.
\bibitem[{Gomes et~al.(2008)Gomes, Kautz, Sabharwal, and
  Selman}]{sat_solvers_08}
\bibinfo{author}{C.~P. Gomes}, \bibinfo{author}{H.~Kautz},
  \bibinfo{author}{A.~Sabharwal}, \bibinfo{author}{B.~Selman},
\newblock \bibinfo{title}{{Satisfiability Solvers}},
\newblock in: \bibinfo{booktitle}{Handbook of Knowledge Representation},
  volume~\bibinfo{volume}{3} of \textit{\bibinfo{series}{Foundations of
  Artificial Intelligence}}, \bibinfo{publisher}{Elsevier},
  \bibinfo{year}{2008}, pp. \bibinfo{pages}{89--134}.
\bibitem[{Gelain et~al.(2010)Gelain, Pini, Rossi, Venable, and
  Wilson}]{intervalConstr10}
\bibinfo{author}{M.~Gelain}, \bibinfo{author}{M.~Pini},
  \bibinfo{author}{F.~Rossi}, \bibinfo{author}{K.~Venable},
  \bibinfo{author}{N.~Wilson},
\newblock \bibinfo{title}{{Interval-valued soft constraint problems}},
\newblock \bibinfo{journal}{Annals of Mathematics and Artificial Intelligence}
  \bibinfo{volume}{58} (\bibinfo{year}{2010}) \bibinfo{pages}{261--298}.

\end{thebibliography}
